%% file: arxiv.tex
\documentclass[10pt,twocolumn,letterpaper]{article}

\usepackage[pagenumbers]{wacv}      

\usepackage{graphicx}
\usepackage{amsmath}
\usepackage{amssymb}
\usepackage{booktabs}
\usepackage{multirow}
\usepackage{tabularx}
\usepackage{diagbox}
\usepackage{gensymb}
\usepackage[accsupp]{axessibility}
\usepackage[table,xcdraw]{xcolor}

\newcolumntype{Y}{>{\centering\arraybackslash}X}

\usepackage[pagebackref,breaklinks,colorlinks]{hyperref}

\usepackage[capitalize]{cleveref}
\crefname{section}{Sec.}{Secs.}
\Crefname{section}{Section}{Sections}
\Crefname{table}{Table}{Tables}
\crefname{table}{Tab.}{Tabs.}

\def\wacvPaperID{308} 
\def\confName{WACV}
\def\confYear{2024}

\begin{document}

\title{SyntheWorld: A Large-Scale Synthetic Dataset for Land Cover Mapping and Building Change Detection}

\author{Jian Song$^{1,2}$, Hongruixuan Chen$^{1,2}$, and Naoto Yokoya$^{1,2}$\\
$^{1}$The University of Tokyo, Japan\\
$^{2}$RIKEN AIP, Japan\\
{\tt\small song@ms.k.u-tokyo.ac.jp, Qschrx@gmail.com, yokoya@k.u-tokyo.ac.jp}
}

\maketitle
\begin{center}
\begin{figure*}[htbp]
  \centering
  \includegraphics[width=\textwidth]{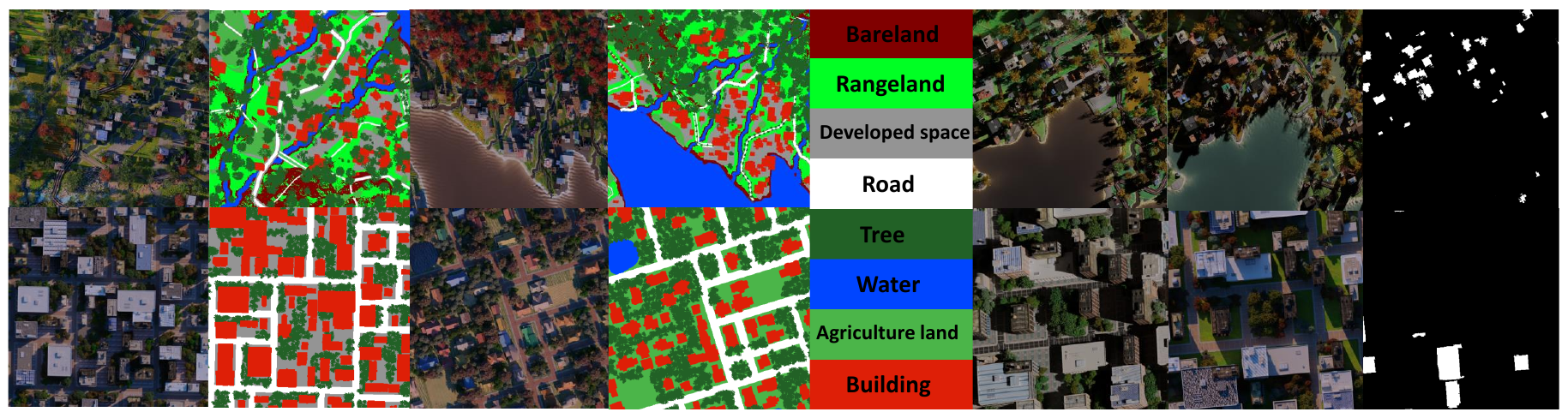}
  \caption{Examples of SyntheWorld dataset.}
  \label{fig:sample_figure}
\end{figure*}
\end{center}

\begin{abstract}
   Synthetic datasets, recognized for their cost effectiveness, play a pivotal role in advancing computer vision tasks and techniques. However, when it comes to remote sensing image processing, the creation of synthetic datasets becomes challenging due to the demand for larger-scale and more diverse 3D models. This complexity is compounded by the difficulties associated with real remote sensing datasets, including limited data acquisition and high annotation costs, which amplifies the need for high-quality synthetic alternatives. To address this, we present SyntheWorld, a synthetic dataset unparalleled in quality, diversity, and scale. It includes 40,000 images with submeter-level pixels and fine-grained land cover annotations of eight categories, and it also provides 40,000 pairs of bitemporal image pairs with building change annotations for building change detection. We conduct experiments on multiple benchmark remote sensing datasets to verify the effectiveness of SyntheWorld and to investigate the conditions under which our synthetic data yield advantages. The dataset is available at \url{https://github.com/JTRNEO/SyntheWorld}.
\end{abstract}

\section{Introduction}
\label{sec:intro}
High-resolution remote sensing image processing is vital for urban planning, disaster response, and environmental monitoring. Although advances in deep neural networks and the emergence of various benchmark datasets have led to significant progress in these research areas, the unique aspects of remote sensing image processing tasks still present many challenges.

First, acquiring large-scale datasets that compare with those in computer vision and natural language processing is difficult due to the sensitivity, privacy, and commercial considerations of remote sensing data. As a result, remote sensing datasets tend to be significantly smaller. Second, compared to fields like computer vision or natural language processing, remote sensing data annotation is both more costly and time-intensive. For example, annotating a $1024\times1024$ image from a large land cover mapping dataset such as~\cite{xia2023openearthmap} usually takes more than two hours. Finally, variations in image capture conditions such as sensor type, image acquisition season, and geographical location introduce a severe domain shift problem in remote sensing image processing. 

Synthetic datasets, with their low-cost acquisition, high fidelity, and diversity, present a viable solution to these challenges. In the field of computer vision, numerous high-quality synthetic datasets~\cite{butler2012naturalistic,mayer2016large,ros2016synthia,richter2016playing,handa2016scenenet,wrenninge2018synscapes} have already emerged, primarily serving tasks such as semantic segmentation, depth estimation, optical flow estimation, and 3D reconstruction of street-view and indoor-view scenario. However, high-quality synthetic datasets for remote sensing are scarce in comparison. The most important reason is, as described in~\cite{kong2020synthinel}, in a virtual world constructed for street-view or indoor-view scenes, the distance between the sensor and the target location is relatively small (a few or tens of meters), with the main focus being on pedestrians, vehicles, road signs, or various furniture, resulting in a relatively small virtual world size. In contrast, in remote sensing scenarios, sensors are often located tens of thousands of meters away from the target virtual world, making even a relatively small virtual world extend over several square kilometers, while maintaining a multitude of diverse targets, such as thousands of trees in different poses and hundreds of buildings of different styles. This makes the construction of large-scale synthetic remote sensing datasets exceptionally challenging.

Upon a thorough survey of the available synthetic remote sensing datasets~\cite{bourdis2011constrained,zou2020game,kong2020synthinel,shermeyer2021rareplanes,xu2022simpl,reyes2022syntcities}, we discern that each of them has specific limitations. First, most existing works focus on a single task, such as building segmentation~\cite{zou2020game,kong2020synthinel} or object detection~\cite{shermeyer2021rareplanes,xu2022simpl}. However, there is a notable lack of effective synthetic datasets for critical tasks like multi-class land cover mapping and building change detection. Furthermore, these datasets exhibit limited diversity due to constraints associated with the size of the virtual world and the tools used. They either emulate real-world cities to create a limited number of virtual environments or use real remote sensing images as the background. Furthermore, when it comes to 3D models in the virtual world, existing methodologies consistently rely on predefined textures, layouts, and geometries, resulting in a restrictive range of styles for buildings, trees, and other land objects.

In this work, we use the freely available open-source 3D modeling software Blender~\cite{blender}, along with various plugins from the Blender community, GPT-4~\cite{openai2023gpt4}, and the Stable Diffusion model~\cite{rombach2022high}, to develop a procedural modeling system specifically for generating high-resolution remote sensing datasets. We present SyntheWorld, the largest high-resolution remote sensing image dataset for land cover mapping and building change detection tasks. \cref{fig:sample_figure} displays some examples from the proposed SyntheWorld dataset.

The main contributions of this work are:
\begin{itemize}
\item We introduce SyntheWorld, the first fully synthetic high-resolution remote sensing dataset, which integrates procedural 3D modeling techniques with Artificial Intelligence Generated Content (AIGC).
\item We use SyntheWorld as the first synthetic dataset specifically designed to improve performance in two crucial tasks: multi-class land cover mapping and building change detection.
\item We propose the Relative Distance Ratio (RDR), a new metric designed to quantify the conditions under which the synthetic dataset can drive performance improvements.
\item Through comprehensive experiments on various remote sensing benchmark datasets, we demonstrate the utility and effectiveness of our dataset.
\end{itemize}

\section{Related Works}
\subsection{Remote Sensing Image Processing Tasks}
\subsubsection{Land Cover Mapping}
The discipline of land cover mapping is a crucial component of remote sensing image processing, where the goal is to categorize and depict physical features on Earth's surface, such as grass, trees, water bodies, bareland, buildings, etc. This task resembles semantic segmentation in traditional computer vision. Although the introduction of benchmark datasets for real-world scenarios, such as DeepGlobe~\cite{demir2018deepglobe}, LoveDA~\cite{wang2021loveda}, and OpenEarthMap (OEM)~\cite{xia2023openearthmap}, has made significant advances in associated research, there is still a clear need for high-quality synthetic datasets. This is an area where the field of computer vision has made significant progress. Recognizing this gap, we were motivated to create SyntheWorld, a synthetic dataset crafted to improve performance in land cover mapping tasks.
\subsubsection{Building Change Detection}
The task of building change detection forms another crucial component within the realm of remote sensing image processing. It involves the identification and localization of modifications in man-made structures, especially buildings, over time, achieved through the analysis of images of the same area captured at different intervals. It is an indispensable technique for assessing damage in scenarios such as earthquakes, hurricanes, or floods, and for monitoring urban development and expansion over time. Typical annotations for this task involve binary masks, with networks trained to predict areas of building change based on input image pairs from two time points. While the emergence of benchmark real-world datasets such as WHU-CD~\cite{ji2018fully}, LEVIR-CD+~\cite{chen2020spatial}, and SECOND~\cite{yang2020semantic} have provided the field with valuable data resources, the lack of high-quality synthetic datasets has hindered the pace of related research. 

\subsection{Existing Synthetic Datasets}
\subsubsection{Street-view \& Indoor-view}
As we mentioned, the availability of large, high-quality synthetic datasets for street-view and indoor-view has driven the development of related techniques in traditional computer vision. The MPI Sintel Dataset~\cite{butler2012naturalistic} is widely used for training and evaluating optical flow algorithms, capturing natural scenes and motions in its synthetic dataset derived from an animated film. SceneFlow~\cite{mayer2016large}, with more than 35,000 synthetic stereo video sequences, is designed for the evaluation of optical flow, disparity, and scene flow algorithms. SYNTHIA~\cite{ros2016synthia}, a dataset composed of 9,400 multi-viewpoint frames from a virtual city, targets urban scene understanding tasks with its pixel-level semantic annotations. The GTA5 dataset~\cite{richter2016playing}, comprising 24,966 synthetic images from the perspective of a car in virtual cities, is tailored to the understanding of urban scenes with its pixel-level semantic annotations compatible with the Cityscapes dataset~\cite{Cordts2016Cityscapes}. Synscapes~\cite{wrenninge2018synscapes}, featuring 25,000 photorealistic street scenes, aims to improve the performance of computer vision models in outdoor scenes with its precise semantic labels. Finally, SceneNet~\cite{handa2016scenenet}, a diverse synthetic dataset of over 5 million indoor scenes with RGB-D images and semantic labels, is designed for indoor scene understanding tasks.

\subsubsection{Overhead-view}
The AICD dataset~\cite{bourdis2011constrained}, one of the earliest datasets with an overhead view, uses the Virtual Battle Station 2 game engine to simulate building alterations. Despite its 1,000 pairs of $800\times600$ RGB image pairs with building change masks, its 500 change instances are limited compared to the tens of thousands found in real-world datasets. The GTA-V-SID dataset~\cite{zou2020game}, extracted from the GTA-V game, covers a 100$km^2$ area with 121 $500\times500$ aerial RGB images. Although it is useful for building segmentation tasks, its 1m GSD limits performance in high-resolution remote sensing datasets. Syntinel-1~\cite{kong2020synthinel}, the first high-resolution synthetic remote sensing dataset for building segmentation, is based on CityEngine and offers a variety of urban styles. The Syntcities dataset~\cite{reyes2022syntcities} is for disparity estimation in remote sensing images, featuring three virtual cities and 8,100 pairs of high-resolution images. RarePlanes~\cite{shermeyer2021rareplanes}, a semi-synthetic dataset for aircraft object detection, combines real WorldView-3 satellite imagery and 3D models. 

\section{Dataset Generation and Description}
Constructing a virtual city manually is time-consuming. Comparatively, SyntheWorld differs from existing overhead-view synthetic datasets by using procedural modeling. 
Previous studies in computer graphics have explored procedural modeling for cities and buildings~\cite{musgrave1989synthesis,muller2006procedural,kim2018procedural}, but none have utilized these techniques for the creation of overhead view datasets. We create our own procedural rules to create 3D geometries and apply textures derived from generative models, which minimize labor costs and enrich diversity. 
\begin{figure}[t!]
\centering
\includegraphics[width=\linewidth]{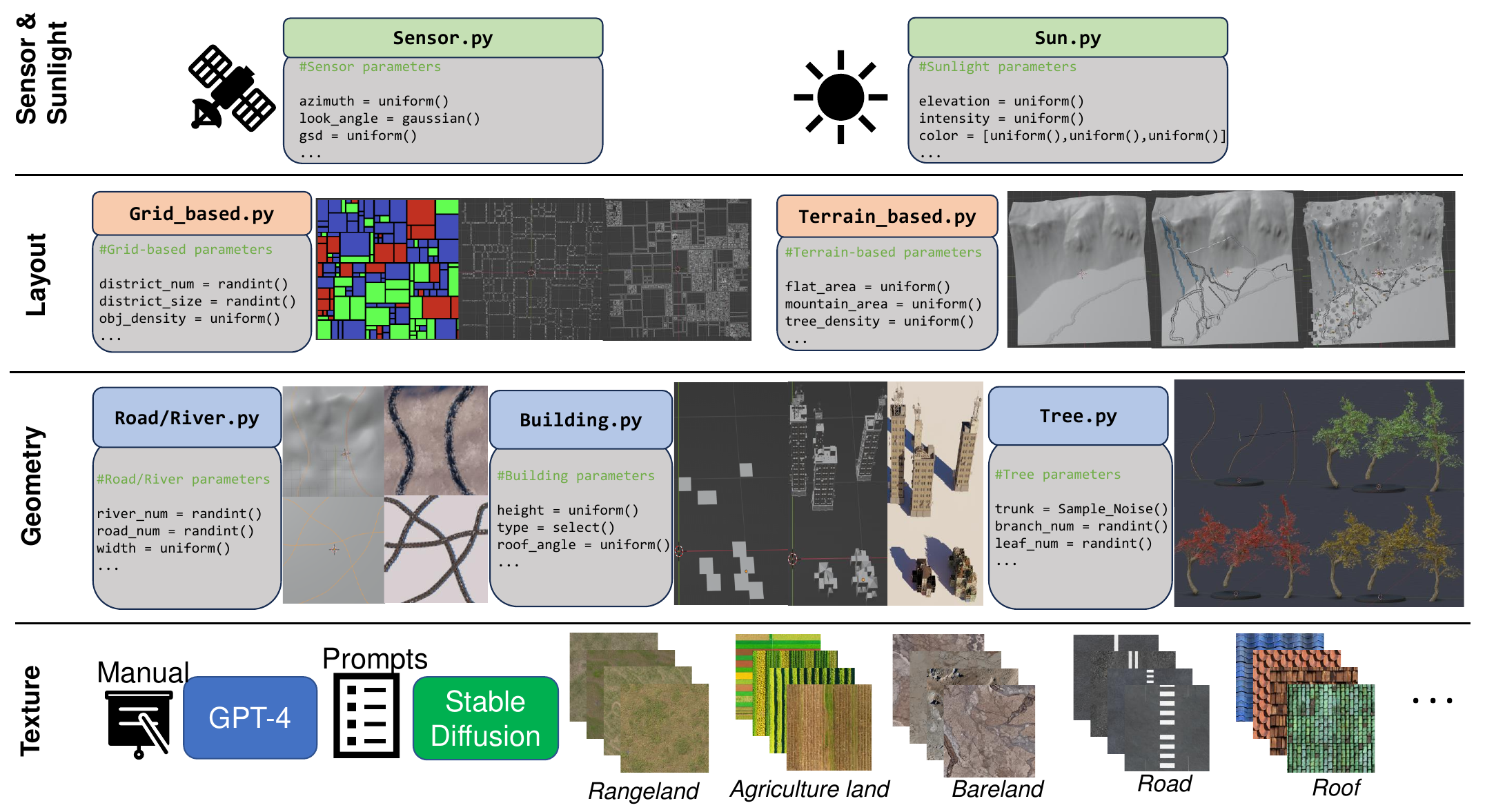}
\caption{The essential components for building SyntheWorld dataset.}
\label{fig:components}
\end{figure}

\begin{table*}[htbp]
\scriptsize
\centering
\setlength{\tabcolsep}{5pt}
\begin{tabularx}{\textwidth}{@{}l*{7}{>{\centering\arraybackslash}X}@{}}
\toprule
\multirow{2}{*}{RS Synthetic Datasets} & \multicolumn{7}{c}{Features and Composition} \\
\cmidrule(lr){2-8} 
& GSD (m) & Task & $\#$ of images & Image Size & Automatic Labeling & Fully Synthetic & Procedural Modeling \\
\midrule
AICD~\cite{bourdis2011constrained} & $-$ & BCD & $1,000$ pairs & $800\times600$ & $\surd$ & $\surd$ & $\times$\\
GTA-V-SID~\cite{zou2020game} & $1$ & BS & $121$ & $500\times500$ & $\times$ & $\surd$ & $\times$ \\
Synthinel-1~\cite{kong2020synthinel} & $0.3$ & BS & $1,054$ & $572\times572$ & $\surd$ & $\times$ & $\times$ \\
RarePlanes~\cite{shermeyer2021rareplanes} & $0.31 \sim 0.39$ & OD & $50,000$ & $512\times512$ & $\surd$ & $\times$ & $\times$ \\
SyntCities~\cite{reyes2022syntcities} & $0.1, 0.3, 1.0$ & DE & $8,100$ pairs & $1024\times1024$ & $\surd$ & $\times$ & $\times$ \\
\midrule
\multirow{2}{*}{SyntheWorld (Ours)} & $0.3 \sim 0.6$ & \multirow{2}{*}{BS/LC/BCD} & $30,000$ pairs & $512\times512$ & \multirow{2}{*}{$\surd$} & \multirow{2}{*}{$\surd$} & \multirow{2}{*}{$\surd$} \\
& $0.6 \sim 1.0$ & & $10,000$ pairs & $1024\times1024$ & & & \\
\bottomrule
\end{tabularx}
\caption{Features and composition comparison among remote sensing synthetic datasets. LC: land cover mapping. BCD: building change detection. BS: building segmentation. OD: object detection. DE: disparity estimation.}
\label{tab:feature comparison}
\end{table*}

\subsection{Generation Workflow}
\textbf{Layout}. We adopt grid-based and terrain-based methods for the virtual world, as illustrated in~\cref{fig:components}. For the grid-based method, we randomly slice a grid of 0.25-0.36$km^2$ into several blocks of varying numbers and sizes, placing different types of buildings and trees in each block, and the boundaries between the blocks serve as our road system. It mainly simulates the more regular city and suburban layouts, and also contributes to the production of 0.3-0.6m GSD synthetic remote sensing images. For the terrain-based method, we use random noise textures to generate terrains such as mountains, plains, and oceans with ranges of 1-2$km^2$. Placing rivers, roads, buildings, and trees according to carefully designed rules based on \emph{Geometry Nodes} in Blender~\cite{blender}, this method mimics irregular layouts in developing regions. It mainly contributes to the production of 0.6-1.0m GSD synthetic remote sensing images.

\textbf{Geometry}. The geometry row in~\cref{fig:components} demonstrates our approach to procedurally model trees and buildings. For buildings, we use random noise to cut out differently shaped grids on a flat plane, which we then extrude into 3D geometries following pre-set rules. Users can control predefined parameters to generate an infinite number of different geometric styles. We distribute predefined asset components (walls, roofs, windows, etc.) to the geometry and finally map the texture generated by AIGC to the building. For trees, we use random-shaped curves as trunks and distribute different styles of tree components to the curve following certain rules.

\textbf{Texture}. The last row in~\cref{fig:components} shows examples of our process for generating corresponding texture assets using AIGC. In terms of operational specifics, we first make a Stable Diffusion usage guide as a prompt to help GPT-4 understand its workings and prompt forms. We then provide excellent prompts as examples and ask GPT-4 to generate different themed prompts for different types of textures. In total, we generated around 140,000 seamless textures for different geometry to build SyntheWorld, far exceeding the number of textures used by existing overhead-view datasets. See the supplementary material for detailed prompts and generated images.
\subsection{Structure of Dataset}
As shown in~\cref{tab:feature comparison}, SyntheWorld is a comprehensive image dataset, consisting of 40,000 pairs of images. Of these, 30,000 pairs have a GSD ranging from 0.3 to 0.6 m, with each image having size $512\times512$. The remaining 10,000 pairs have a GSD of 0.6 to 1.0 m and a larger image size of $1024\times1024$.

Each pair in the dataset contains a post-event image, which is utilized for the land cover mapping task. These post-event images are accompanied by semantic labels of eight categories, as shown in~\cref{fig:sample_figure}. These categories are consistent with those of the OEM~\cite{xia2023openearthmap} dataset. Correspondingly, the pre-event images are derived by introducing variability in each scene. This involves different textures, lighting parameters, and camera settings. Additionally, there is a 10$\%$ to 50$\%$ chance that any given building in the scene might be removed.

Both pre-event and post-event images from each pair are used collectively for the building change detection task. Accordingly, the dataset comes with 40,000 binary classification masks corresponding to this task.

The off-nadir angle of all images ranges from $-25\degree$ to $25\degree$ and follows a Gaussian distribution with a certain mean $0\degree$ and variance $2.3\degree$. Similarly, we simulate the sun's position during the day in most countries by adjusting the zenith (ranging between $25\degree$ to $35\degree$) and the elevation parameters (ranging between $45\degree$ to $135\degree$), as guided by the documentation of the Pro Atom~\cite{ProAtmo} addon in the Blender community, both parameters following a uniform distribution. This inclusion of various viewing angles and sun elevation enhances the robustness of SyntheWorld and ensures its applicability to a wide range of real-world datasets.

\subsection{Comparison with Existing Synthetic Datasets}

As depicted in \cref{tab:feature comparison}, we provide a comparative analysis of SyntheWorld and existing synthetic remote sensing datasets, in terms of their features and composition. The Task column presents the primary tasks illustrated in the corresponding dataset's literature. 

Regarding label generation, the GTA-V-SID dataset~\cite{zou2020game} consists of screenshots of the GTA-5 commercial video game, with buildings manually annotated. On the contrary, the remaining datasets are capable of automatically generating annotations via the corresponding 3D software.

In terms of complete synthesis, only SyntheWorld achieves this feat. The other datasets have adopted real remote sensing images to some extent as texture or as part of the dataset during their construction.

Finally, in SyntheWorld, most 3D models are generated using procedural modeling, while in other synthetic datasets, the geometric structure and texture of the models are either predefined or meticulously designed by 3D artists. This unique characteristic of SyntheWorld significantly enhances its diversity.

\section{Experiments}
\subsection{Real-world Benchmark Datasets}
To validate the versatility and effectiveness of SyntheWorld, we performed experiments using several high-resolution remote sensing datasets from various real-world scenarios. In the subsequent discussion, we present an in-depth overview of these datasets. In the experiments showcased in this section, we employ ``w'' to signify the utilization of the SyntheWorld dataset and ``w/o'' to indicate its non-use.

For the building segmentation task, we relied on OEM~\cite{xia2023openearthmap} and LoveDA~\cite{wang2021loveda} datasets, as well as INRIA~\cite{maggiori2017can} and BANDON~\cite{pang2023detecting} datasets. The INRIA dataset, which targets building footprint segmentation, incorporates aerial images from ten cities in the United States and Europe at a resolution of 0.3 m. The BANDON dataset stands out with significant off-nadir angles and focuses on urban areas with skyscrapers. It offers high-resolution 0.6m remote sensing images from Beijing and Shanghai.

We turned to OEM and LoveDA datasets again for the multi-class land cover mapping task. The OEM dataset, encompassing 97 regions across 44 countries worldwide, provides high-resolution images with detailed eight-class land cover annotations. The LoveDA dataset offers 0.3m GSD remote sensing images from three diverse regions in China, labeled with seven land cover categories.

In the building change detection task, we harnessed the WHU-CD~\cite{ji2018fully}, LEVIR-CD+~\cite{chen2020spatial}, and SECOND~\cite{yang2020semantic} datasets. The LEVIR-CD+ dataset consists of 987 image pairs, with 637 pairs in the training set and 348 pairs in the test set. SECOND, a semantic change detection dataset, collects 4662 pairs of aerial images from various platforms and sensors across cities like Hangzhou, Chengdu, and Shanghai. The WHU-CD dataset consists of two pairs of super-high-resolution (0.075m) aerial images. We cropped these large training ($21243\times15354$) and testing ($11265\times15354$) images into non-overlapping $512\times512$ patches for our experiments.

\subsection{Building Segmentation}
To compare with existing overhead-view synthetic datasets, which mainly include semantic labels of buildings, we performed building segmentation experiments. We use the DeepLabv3 +~\cite{chen2018encoder} network equipped with ResNet-50~\cite{he2016deep} backbone. We adopted the SGD optimizer~\cite{robbins1951stochastic} for all synthetic datasets, employing a learning rate of 1e-3, a weight decay of 5e-4, and a momentum of 0.9; for the OEM dataset, we opted for a higher learning rate of 1e-2.
\begin{table}[t!]
\centering
{\scriptsize
    \begin{tabular}{lcccc}
        \toprule
        \diagbox{Train on}{Test on} & OEM* & LoveDA* & INRIA & BANDON \\
        \midrule
        GTA-V-SID~\cite{zou2020game} & 2.43 & 0.88 & 1.74 & 1.64 \\
        Synthinel-1~\cite{kong2020synthinel} & 35.37 & 14.13 & 39.89 & 28.19 \\
        SyntCities~\cite{reyes2022syntcities} & 23.61 & 21.39 & 30.39 & 30.01 \\
        SyntheWorld & \textbf{49.26} & \textbf{37.28} & \textbf{45.76} & \textbf{34.01} \\
        \midrule
        OEM*~\cite{xia2023openearthmap} & 80.48 & 55.35 & 75.61 & 64.19 \\
        \bottomrule
    \end{tabular}
}    
    \caption{mIoU(\%) results of the building segmentation task using DeepLabv3+. * means to use only the part of the building label in the dataset.}
    \label{tab:bstable}
\end{table}

The results are presented in~\cref{tab:bstable}. The GTA-V-SID~\cite{zou2020game} dataset underperforms on various high-resolution real-world datasets due to its smaller quantity and 1m GSD. The model trained on the SyntheWorld dataset outperforms other datasets on four real-world datasets, especially on the OEM and LoveDA datasets. These two datasets include a considerable number of buildings in developing or developed areas. Thus, the performance of SyntheWorld far exceeds that of other competitors in these two datasets.
As the buildings in the INRIA~\cite{maggiori2017can} and BANDON~\cite{pang2023detecting} datasets are predominantly high-rises in urban areas or well-organized detached houses in suburban areas, the advantage of the SyntheWorld dataset is not as evident as in the other two datasets, but still shows the best performance. Furthermore, the last column of~\cref{tab:bstable} shows the performance of the model trained on the OEM dataset and tested on other datasets. Although SyntheWorld significantly outperforms other synthetic datasets, there is still a gap compared to real-world datasets.

In~\cref{fig:features} (a), we also visualized the feature extract of the well-trained ResNet-50 of all synthetic and real datasets using UMAP~\cite{mcinnes2018umap}. In terms of feature space, SyntheWorld is closer to real-world datasets than any existing synthetic datasets.

\subsection{Land Cover Mapping}
SyntheWorld is the first synthetic dataset that offers consistent annotations compatible with high-resolution real-world benchmarks. In this section, we primarily discuss the performance of SyntheWorld in the land cover mapping task.
\subsubsection{Cross-dataset Experiments}
\label{sec:cross-datasets}
To evaluate the enhancements brought about by using SyntheWorld, we adopted the mixed training strategy~\cite{richter2016playing} often used with synthetic datasets, a batch size of 8, including 7 real images and 1 synthetic image per batch. The model was trained using DeepLabv3+ with the SGD optimizer and an initial learning rate of 1e-2, accompanied by a weight decay of 5e-4, and a momentum of 0.9. All experiments were trained for 100 epochs on a Tesla A100 GPU. Specifically, we map the rangeland class and the developed space class in OEM and SyntheWorld to the background class in LoveDA to keep the classes consistent.
\begin{table}[t!]
\centering
{\scriptsize
    \begin{tabular}{lcc}
        \toprule
        Datasets & w/o & w/ \\
        \midrule
        OEM~\cite{xia2023openearthmap} & \textbf{66.96} & 66.84 \\
        LoveDA~\cite{wang2021loveda} & 51.14 & \textbf{53.32} \\
        O→L & \textbf{35.28} & 34.83 \\
        L→O & 21.95 & \textbf{25.24} \\
        \bottomrule
    \end{tabular}
}
    \caption{Land cover mapping mIoU(\%) outcomes from intra-dataset and cross-dataset evaluations, utilizing the DeepLabv3+ model for all experiments. O→L denotes training on the OEM training set and testing on the LoveDA validation set, while L→O represents the converse.}
    \label{tab:lcratiotable2}
\end{table}

~\cref{tab:lcratiotable2} outlines the results obtained by integrating training images from a real-world dataset with SyntheWorld and the results of cross-dataset tests using the SyntheWorld dataset. Incorporating SyntheWorld with the entire OEM~\cite{xia2023openearthmap} training set does not result in performance enhancements. Similarly, combining SyntheWorld with the OEM training set and subsequently testing on LoveDA~\cite{wang2021loveda} slightly reduces model efficacy. However, when we merge SyntheWorld with the LoveDA and test on the same, the model's mIoU increases by 2.18 points. In addition, a 3.29-point improvement in mIoU is observed when testing the OEM test set after integrating SyntheWorld and LoveDA.
\begin{figure}
\centering
\begin{tabular}{cc}
\includegraphics[width=3.87cm]{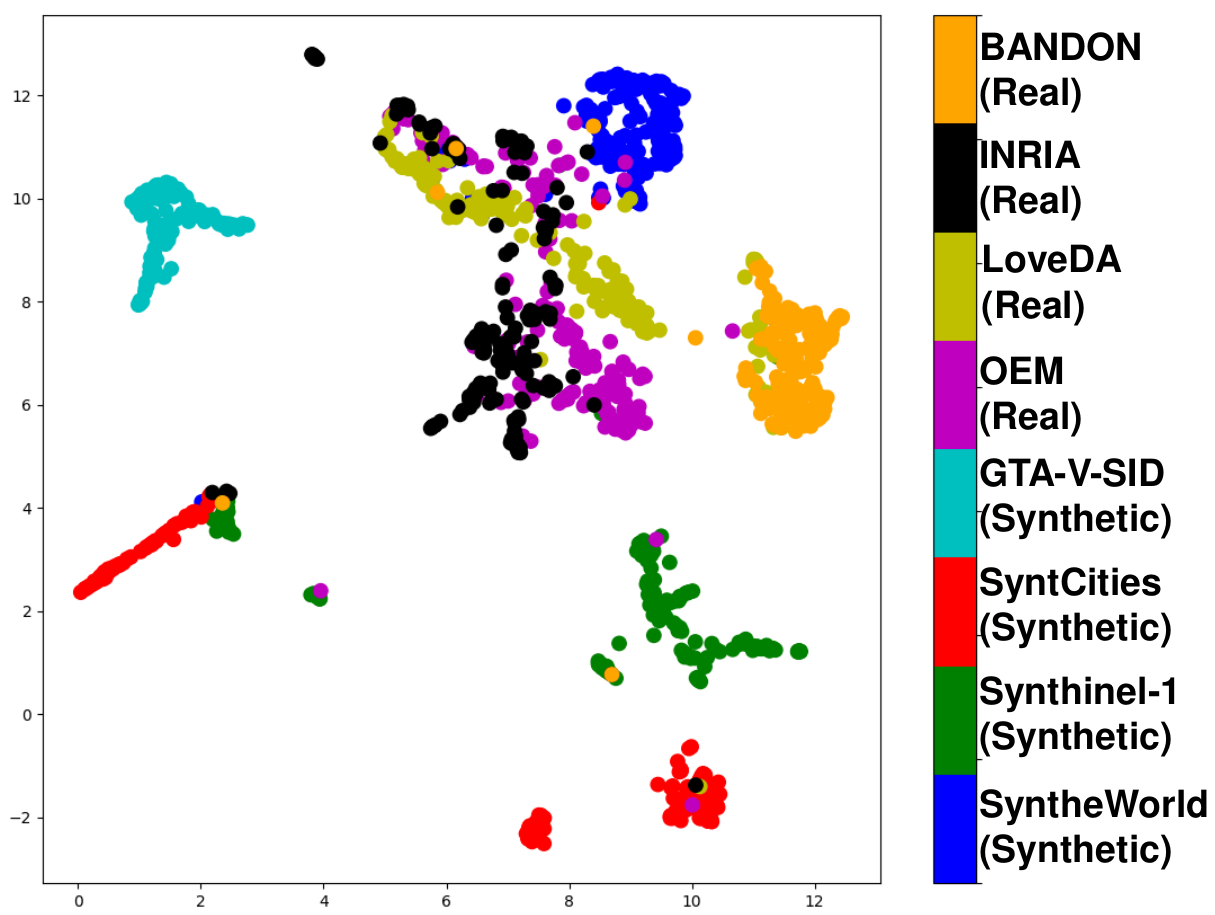}&
\includegraphics[width=3.62cm]{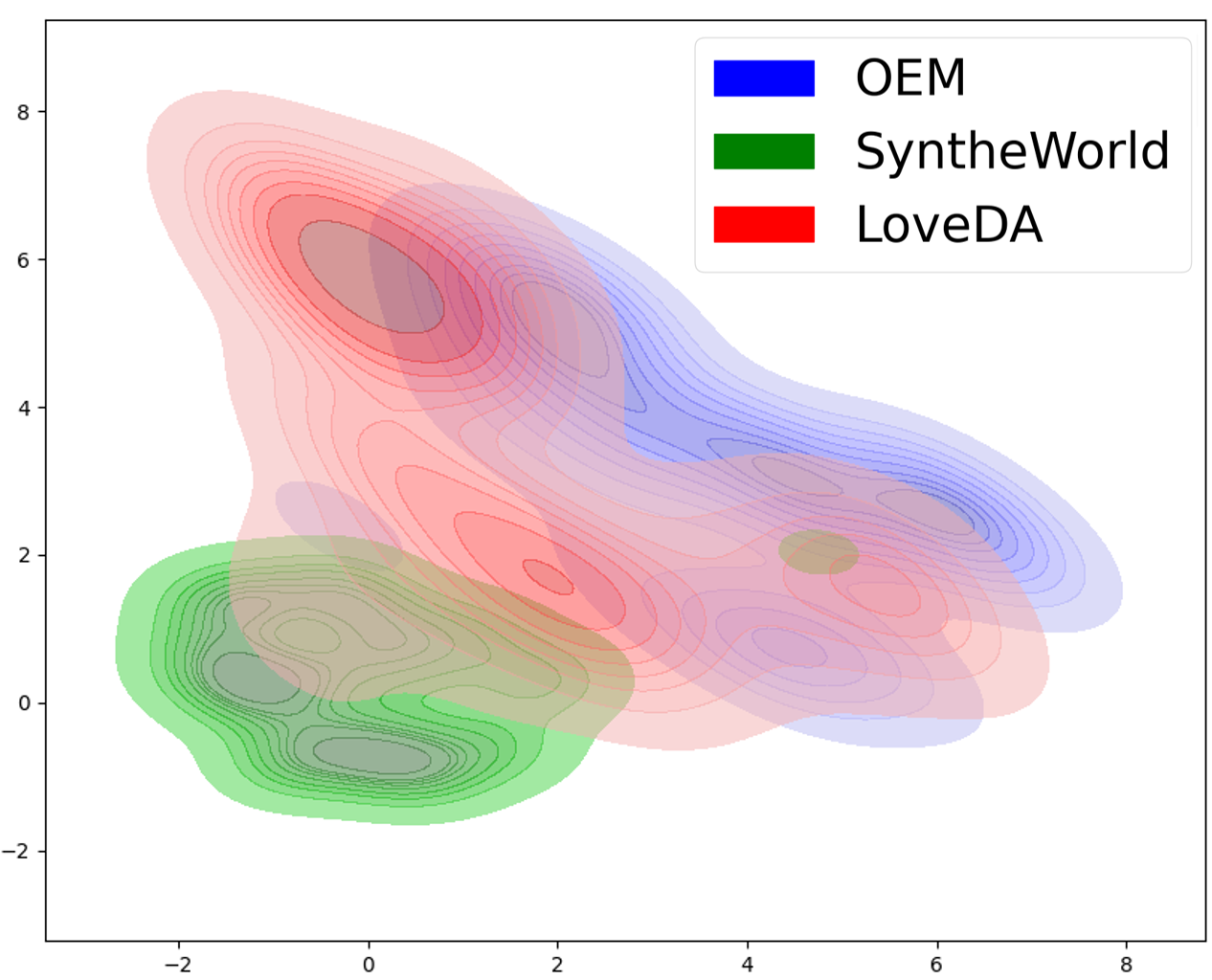}
\\(a)&(b)
\end{tabular}
\caption{(a) 2D UMAP visualization of synthetic and real datasets. We use ResNet-50 pre-trained on the OEM dataset as the feature extractor; (b) Colormap of density estimation for SyntheWorld, OEM, and LoveDA dataset.}
\label{fig:features}
\end{figure}

To investigate the observed phenomenon, we made density estimation maps for the three datasets as displayed in~\cref{fig:features} (b). This reveals a notable overlap between SyntheWorld and OEM, with a lesser overlap in relation to LoveDA. The expansive coverage of the OEM dataset surpasses that of LoveDA and SyntheWorld. This finding sheds light on the patterns observed in~\cref{tab:lcratiotable1}. The vast diversity of the OEM dataset effectively captures the most data diversity inherent in SyntheWorld. Therefore, no performance enhancement results from integrating SyntheWorld. Nevertheless, the substantial overlap between SyntheWorld and OEM enables a performance boost when SyntheWorld is merged with LoveDA and tested on OEM. Conversely, the lesser overlap between SyntheWorld and LoveDA means that integrating SyntheWorld during OEM training does not lead to improvements in the LoveDA test set.
\begin{table}[t!]
\centering
{\scriptsize
    \begin{tabular}{lcccccc}
        \toprule
        Datasets & \multicolumn{2}{c}{1\%} & \multicolumn{2}{c}{5\%} & \multicolumn{2}{c}{10\%} \\
        \cmidrule(r){2-3} \cmidrule(lr){4-5} \cmidrule(l){6-7}
        & w/o & w/ & w/o & w/ & w/o & w/ \\
        \midrule
        OEM~\cite{xia2023openearthmap} & 40.9 & \textbf{45.01} & 52.21 & \textbf{54.0} & 58.40 & \textbf{59.31} \\
        LoveDA~\cite{wang2021loveda} & 34.59 & \textbf{36.75} & 42.38 & \textbf{44.58} & 45.27 & \textbf{48.12} \\
        \bottomrule
    \end{tabular}
}
    \caption{mIoU(\%) results from the DeepLabv3+ model, trained both with and without SyntheWorld, and deployed on two real-world land cover mapping datasets at various proportions of real image utilization.}
    \label{tab:lcratiotable1}
\end{table}

Subsequently, we assessed performance when integrating SyntheWorld with varying proportions of real-world datasets.~\cref{tab:lcratiotable1} presents the findings. Irrespective of the real-world dataset being OEM or LoveDA, the integration of SyntheWorld consistently enhances model performance when the quantity of training data is limited.

\subsubsection{Cross-domain Experiments}
\label{subsec:cross-domain}
In order to examine the performance of SyntheWorld in out-of-domain test scenarios, we partition the OEM~\cite{xia2023openearthmap} dataset into seven distinct continents. Africa, Asia, Europe, Central America, North America, South America, and Oceania. Simultaneously, for the LoveDA~\cite{wang2021loveda} dataset, we conducted experiments using urban and rural areas as separate domains. We conduct experiments with various decoders and encoders; in this section, we show the results of one model. See supplementary material for more results from different models, dataset division, and experimental setup.

\begin{figure*}
\centering
\begin{tabular}{ccc}
\includegraphics[width=4.5cm]{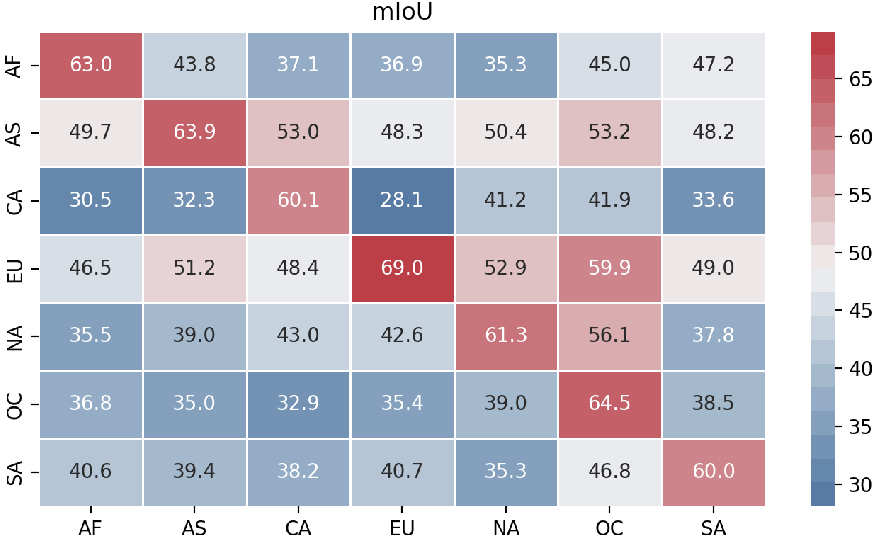}&
\includegraphics[width=4.5cm]{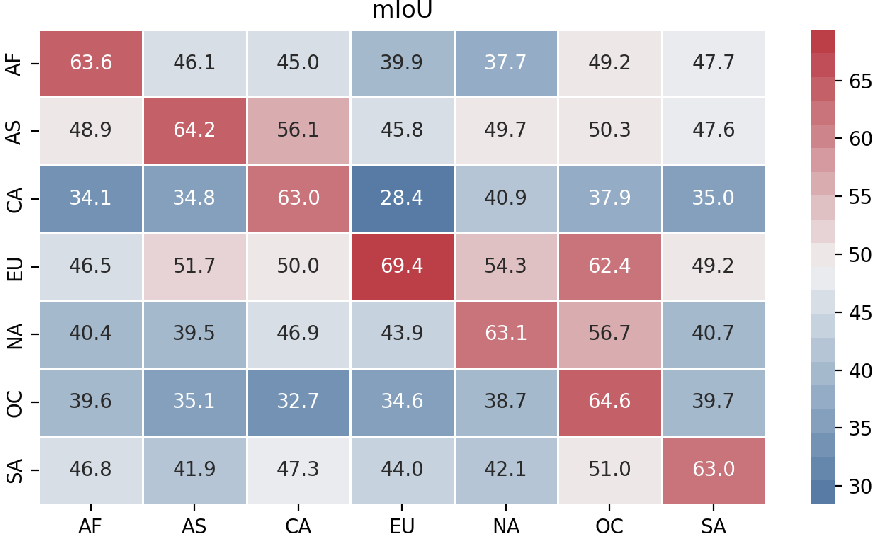}&
\includegraphics[width=4.5cm]{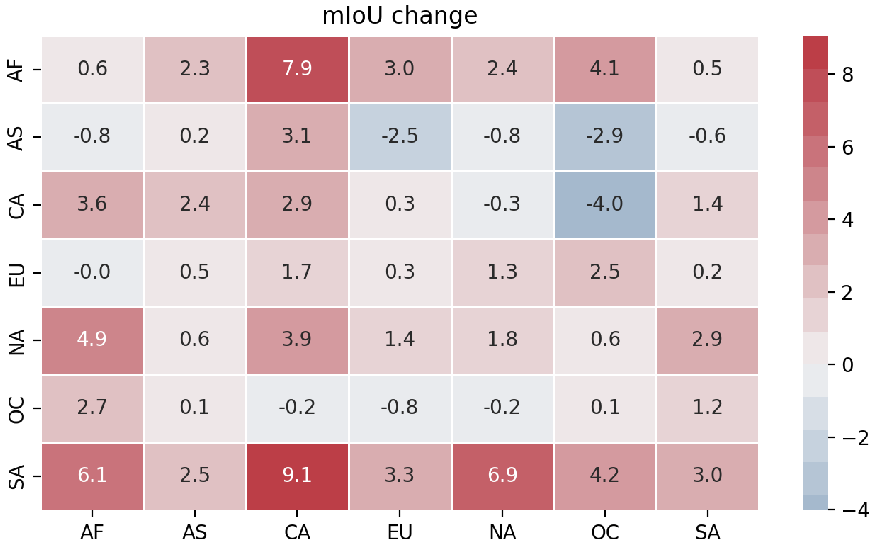}
\\(a)&(b)&(c)
\end{tabular}
\caption{Results of continent-wise in-domain and out-of-domain land cover mapping experiments of OEM dataset. The x-axis represents the target domain and the y-axis represents the source domain. U-Net with EfficientNet-B4 encoder is used for all experiments. (a) The mIoU results of without using SyntheWorld; (b) The mIoU results of mixed training with SyntheWorld; (c) Changes in mIoU.}
\label{fig:continent}
\end{figure*}
\begin{figure}[htbp]
\centering
\includegraphics[width=0.7\linewidth]{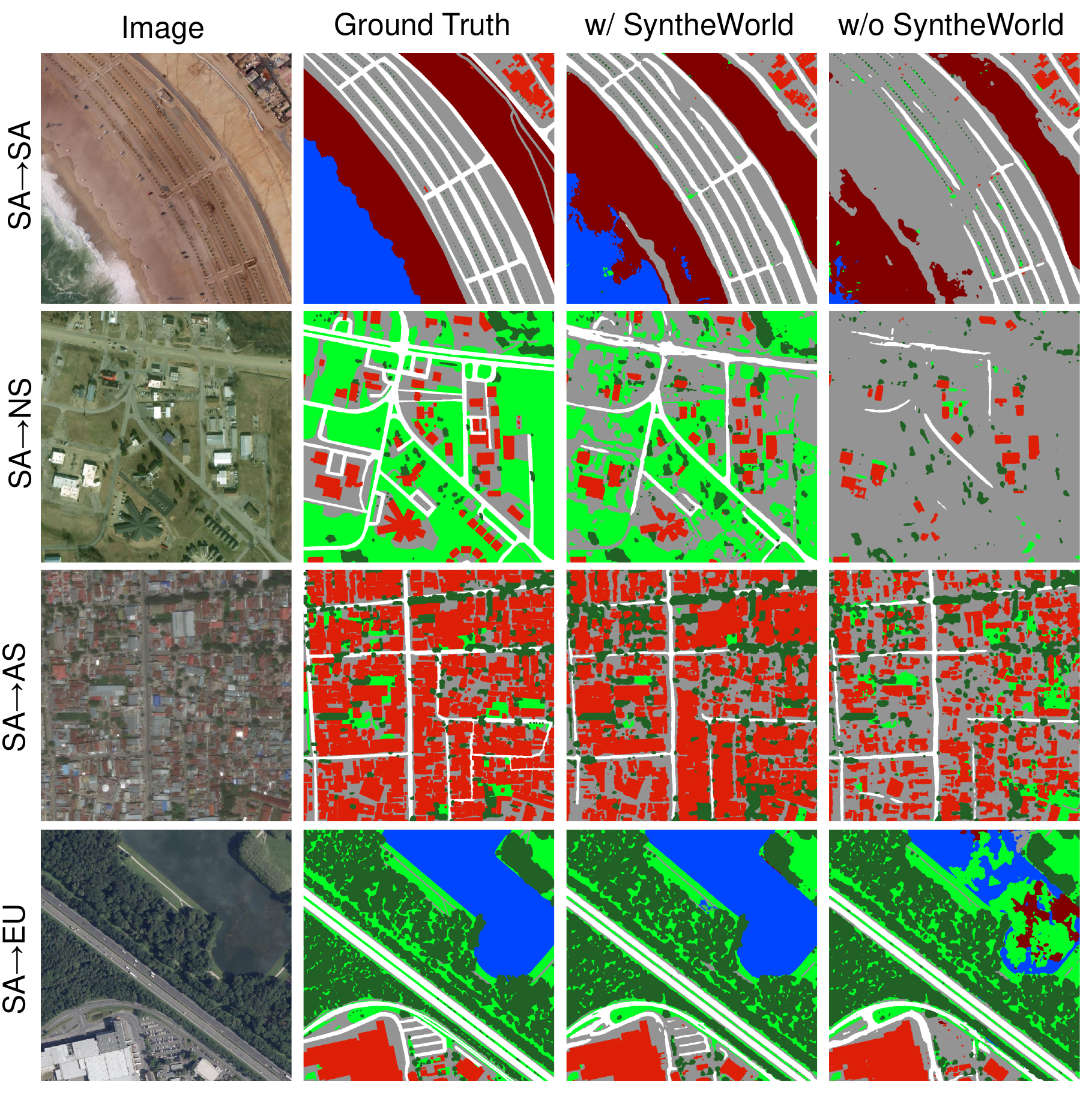}
\caption{Qualitative results by U-Net model of continent-wise land cover mapping task.}
\label{fig:lc_vis}
\end{figure}

\textbf{Continent-wise experimental results}. \cref{fig:continent} displays the results of cross-continent experiments in the OEM dataset using the U-Net~\cite{ronneberger2015u} architecture with the EfficientNet-B4~\cite{tan2019efficientnet} encoder. We can observe that our SyntheWorld dataset can significantly enhance performance across most dataset pairs. Also, we show in~\cref{fig:lc_vis} the qualitative results when synthetic data can lead to a boost. More results can be found in the supplementary material. However, in some cases, the synthetic dataset does not yield a substantial improvement and could even degrade the model performance. It is crucial to investigate the reasons for such enhancement and impairment for the use of synthetic datasets. Therefore, we have conducted a further analysis of these results in~\cref{sec:rdr}.

\begin{table}[t!]
\centering
{\scriptsize
    \begin{tabular}{lcccc}
        \toprule
        \diagbox[width=8em]{Train on}{Test on} & \multicolumn{2}{c}{Urban} & \multicolumn{2}{c}{Rural} \\
        \cmidrule(r){2-3} \cmidrule(l){4-5}
        & w/o & w & w/o & w \\
        \midrule
        Urban & 47.00 & \textbf{50.32} & 33.44 & \textbf{37.95} \\
        Rural & 36.86 & \textbf{38.17} & 48.64 & \textbf{51.66} \\
        \bottomrule
    \end{tabular}
}    
    \caption{Land cover mapping results, measured in mIoU(\%), from cross-domain experiments involving urban and rural areas of the LoveDA dataset.}
    \label{tab:urbanrural}
\end{table}

\textbf{Urban-Rural experimental results}. We conducted similar cross-domain experiments on the LoveDA dataset, which includes two domains, rural and urban. The results are illustrated in~\cref{tab:urbanrural}. We found that the SyntheWorld dataset enhances model performance in both in-domain and out-of-domain tests.

\subsubsection{Relative Distance Ratio}
\label{sec:rdr}
The cross-domain experiments discussed in~\cref{sec:cross-datasets} and~\cref{subsec:cross-domain} show that the SyntheWorld dataset does not always yield significant improvements. This highlights the need to understand the underlying causes. We introduce a metric, the Relative Distance Ratio (RDR), aiming to quantify the relationship between source, target, and synthetic datasets and clarify when synthetic data can bring improvements.

For measuring the distance between datasets, various methods have been discussed in the literature~\cite{heusel2017gans,alvarez2020geometric,shen2018wasserstein}. The most commonly used measure of the distance between synthetic and real datasets is the FID score~\cite{heusel2017gans}. Here we adopt the Fréchet Distance, as the measure of distance between different datasets. Since the Inception model~\cite{szegedy2016rethinking} pre-trained on ImageNet~\cite{deng2009imagenet} is not suitable for remote sensing datasets, we use ResNet-50~\cite{he2016deep} pre-trained on the OEM~\cite{xia2023openearthmap} dataset. The formula to compute the FD between any dataset pair is as follows:
\begin{equation}
    FD(x, y) = ||\mu_x - \mu_y||^2 + \text{Tr}(\Sigma_x + \Sigma_y - 2(\Sigma_x\Sigma_y)^{\frac{1}{2}})
    \label{eq:FD}
\end{equation}
where $\mu_x$ and $\mu_y$ denote the mean feature vectors of datasets $x$ and $y$, respectively, and $\Sigma_x$ and $\Sigma_y$ represent the covariance matrices of the corresponding feature vectors. 

Then we denote the source domain dataset as $S$, the target domain dataset as $T$, SyntheWorld as $G$, and the FD between any two datasets as $\delta(., .)$. Afterwards, the distance between the source domain dataset $S$ and the target domain dataset $T$ can be expressed as:
\begin{equation}
    \delta(f_S, f_T) = FD(f_S, f_T)
    \label{eq:FD_ST}
\end{equation}
Similarly, the distance between the target domain dataset $T$ and the synthetic dataset $G$ can be represented as:
\begin{equation}
    \delta(f_T, f_G) = FD(f_T, f_G)
    \label{eq:FD_TG}
\end{equation}

These $f_T$, $f_S$ and $f_G$ are obtained by applying a ResNet-50 model, pre-trained on the OEM dataset.

Subsequently, we can define the Relative Distance Ratio (RDR), denoted as \( \mathcal{R}(f_S, f_T, f_G) \), to be calculated using the following formula:
\begin{equation}
    \mathcal{R}(f_S, f_T, f_G) = \frac{\delta(f_T , f_G )}{\delta(f_S , f_T )}
    \label{eq:RDR}
\end{equation}
\begin{figure}[t]
\centering
\includegraphics[width=0.8\linewidth]{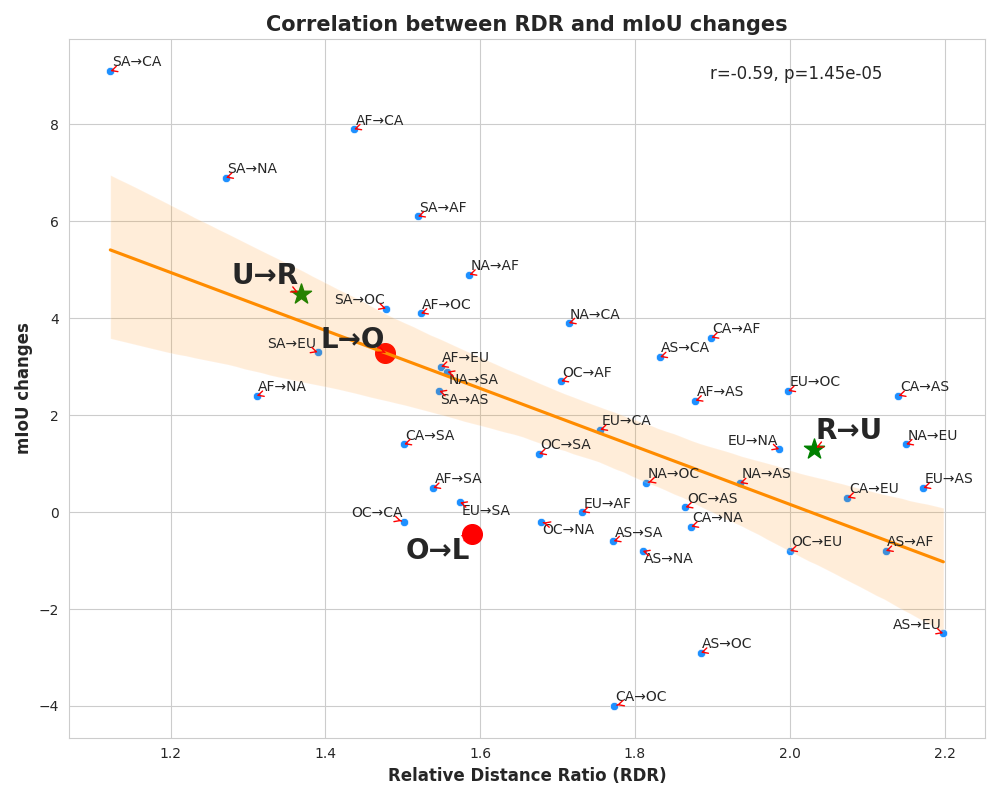}
\caption{Scatter diagram with correlation between mIoU changes and proposed Relative Distance Ratio (RDR).}
\label{fig:cor}
\end{figure}
Intuitively, a smaller $\mathcal{R}$ indicates a greater capacity of the model to integrate knowledge from the synthetic data and transfer it to the target domain. To validate this, we presented a correlation scatter plot in \cref{fig:cor}, which reveals a negative correlation between $\mathcal{R}$ and the improvement in mIoU. This observation aligns with our initial conception of designing the RDR metric. Therefore, the proposed RDR metric effectively serves as a quantitative conditional criterion for employing synthetic data, that is, when $\mathcal{R}$ is large, there is a risk of using synthetic data and vice versa.

\subsection{Building Change Detection}
In this section, we demonstrate the effectiveness of SyntheWorld on the building change detection task. We employ four prevalent building change detection networks, FC-siam-Diff~\cite{daudt2018fully}, STANet-PAM~\cite{chen2020spatial}, DTCDSCN~\cite{liu2020building}, and ChangeFormer~\cite{bandara2022transformer}. We adhere to a mixed training strategy that includes a 7:1 real-to-synthetic image ratio. For ChangeFormer and DTCDSCN we use AdamW~\cite{loshchilov2017decoupled} optimizer with learning rate 1e-4, for the other two models we use Adam optimizer with learning rate 1e-3. Each mixed training experiment is trained for 100 epochs on the Tesla A100 GPU.
\begin{table}[t!]
\centering
{\scriptsize
    \begin{tabular}{lcccccc}
        \toprule
        Datasets & \multicolumn{2}{c}{STANet-PAM} & \multicolumn{2}{c}{DTCDSCN} & \multicolumn{2}{c}{ChangeFormer} \\
        \cmidrule(r){2-3} \cmidrule(lr){4-5} \cmidrule(l){6-7}
        & w/o & w/ & w/o & w/ & w/o & w/ \\
        \midrule
        LEVIR-CD+~\cite{chen2020spatial} & 0.752 & \textbf{0.782} & 0.793 & \textbf{0.812} & 0.784 & \textbf{0.835} \\
        SECOND*~\cite{yang2020semantic} & 0.713 & \textbf{0.733} & 0.712 & \textbf{0.727} & 0.723 & \textbf{0.734} \\
        WHU-CD~\cite{ji2018fully} & 0.707 & \textbf{0.802} & 0.769 & \textbf{0.862} & 0.783 & \textbf{0.836} \\
        \bottomrule
    \end{tabular}
}
    \caption{F1 score resulting from the use or non-use of SyntheWorld across three building change detection benchmark datasets, assessed with three different models.}
    \label{tab:bctable}
\end{table}

\begin{figure}[t!]
\centering
\includegraphics[width=0.7\linewidth]{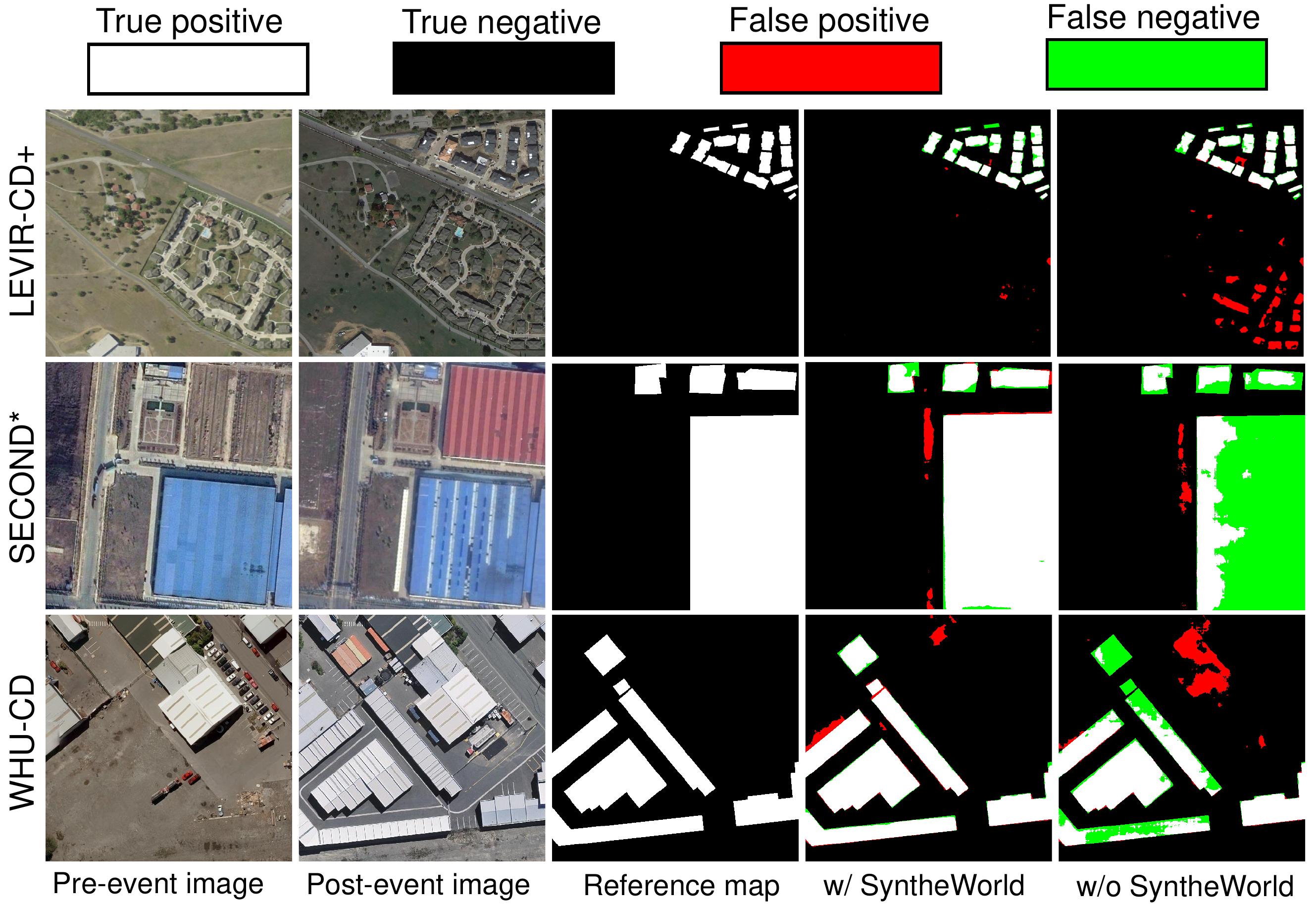}
\caption{Qualitative results by DTCDSCN model of building change detection task on three datasets.}
\label{fig:cd_vis}
\end{figure}

~\cref{tab:bctable} presents the F1 score of three different models applied to three different datasets in the real world. Evidently, for each real-world dataset and each model type, integrating the SyntheWorld dataset induces an improvement, notably for the WHU-CD dataset where it can induce almost a 10-point increase in the F1 score when using the STANet-PAM and DTCDSCN models. Also, we display in~\cref{fig:cd_vis} the qualitative results when using SyntheWorld can lead to enhancements. More results can be found in the supplementary material.
\begin{table}[t!]
\centering
{\scriptsize
    \begin{tabular}{lcccccc}
        \toprule
        \multirow{2}{*}{Datasets} & \multicolumn{2}{c}{1\%} & \multicolumn{2}{c}{5\%} & \multicolumn{2}{c}{10\%} \\
        \cmidrule(r){2-3} \cmidrule(lr){4-5} \cmidrule(l){6-7}
        & w/o & w/ & w/o & w/ & w/o & w/ \\
        \midrule
        LEVIR-CD+~\cite{chen2020spatial} & 0.517 & \textbf{0.646} & 0.636 & \textbf{0.731} & 0.726 & \textbf{0.764} \\
        SECOND*~\cite{yang2020semantic} & 0.401 & \textbf{0.435} & 0.546 & \textbf{0.622} & 0.583 & \textbf{0.631} \\
        WHU-CD~\cite{ji2018fully} & 0.242 & \textbf{0.312} & 0.433 & \textbf{0.638} & 0.510 & \textbf{0.705} \\
        \bottomrule
    \end{tabular}
}    
    \caption{Comparison of F1 scores from the DTCDSCN model trained with and without SyntheWorld, applied on three different real-world datasets at varying ratios of real image use.}
    \label{tab:bcratiotable}
\end{table}

~\cref{tab:bcratiotable} reveals the F1 score of the DTCDSCN model with different proportions of the real-world training set, with and without the incorporation of SyntheWorld. Across all real-world datasets, SyntheWorld invariably provides substantial performance improvement when training data is scarce.
\begin{table}[t!]
\centering
{\scriptsize
    \begin{tabular}{lccc}
        \toprule
        \diagbox[width=10em]{Train on}{Test on} & \textbf{LEVIR-CD+} & \textbf{SECOND*} & \textbf{WHU-CD} \\
        \midrule
        LEVIR-CD+ (Real) & 0.751 & 0.180 & 0.614 \\
        SECOND* (Real) & 0.405 & 0.614 & 0.522 \\
        WHU-CD (Real) & 0.222 & 0.248 & 0.812 \\
        \midrule
        AICD (Synthetic) & 0.094 & 0.267 & 0.092 \\
        SyntheWorld & \textbf{0.419} & \textbf{0.386} & \textbf{0.457} \\
        \bottomrule
    \end{tabular}
}
    \caption{Evaluation of generalizability across multiple building change detection datasets. The table shows the F1 scores. * means to use the part of building change label in SECOND.}
    \label{tab:pure}
\end{table}

~\cref{tab:pure} illustrates the generalizability of the SyntheWorld dataset with the FC-siam-Diff model. We draw comparisons with three real datasets and the AICD synthetic dataset. The results show that by using only the SyntheWorld dataset for training, we can achieve acceptable results on all three datasets. Specifically, compared to the AICD~\cite{bourdis2011constrained} dataset, ours has a significant performance and generalization advantage.

\section{Discussion and Societal Impacts}
We introduced SyntheWorld, the most extensive synthetic remote sensing dataset, used for land cover mapping and building change detection. Its diversity, enhanced by procedural modeling and AIGC, sets it apart from other datasets. Comprehensive experiments validate SyntheWorld's utility and flexibility. Furthermore, we investigate scenarios where SyntheWorld does not enhance performance, proposing the RDR metric for initial exploration of when SyntheWorld can deliver lift.

Notably, SyntheWorld has a significant gap compared to real datasets. This stems from some modeling rules mismatching real-world distributions, a challenge we aim to address in future work. Additional future work involves leveraging SyntheWorld to explore domain adaptation and generalization techniques in remote sensing.

\section*{Acknowledgements}
\begin{flushleft}
This work was supported in part by JST FOREST Grant Number JPMJFR206S; Microsoft Research Asia; and the GSFS Challenging New Area Doctoral Research Grant (Project No. C2303).
\end{flushleft}
{\small
\bibliographystyle{ieee_fullname}
\bibliography{egbib}
}
\clearpage  
\twocolumn[ 
   \centering
   \Large \textbf{Supplementary Material} \\  
   \vspace{1em}  
]
\input{arxiv_sup}

\end{document}

%% file: arxiv_sup.tex
In this Supplementary Material, we will present additional examples from the SyntheWorld dataset, demonstrating its diversity. Following this, we will show our process of directing GPT-4~\cite{openai2023gpt4} to generate effective prompts for Stable Diffusion~\cite{rombach2022high} and its variant models to create land cover textures, including examples of both prompts and generated textures. 

Subsequently, we will provide the dataset divisions and relevant experimental settings for the cross-domain land cover mapping task, along with the quantitative and qualitative results of land cover mapping and building change detection tasks on a broader range of models.

Lastly, we will detail the specific information of all Blender~\cite{blender} addons utilized in the creation of the SyntheWorld dataset.

\section{Expanded Examples from SyntheWorld}
\begin{figure*}[t!]
\centering
\includegraphics[width=0.8\linewidth]{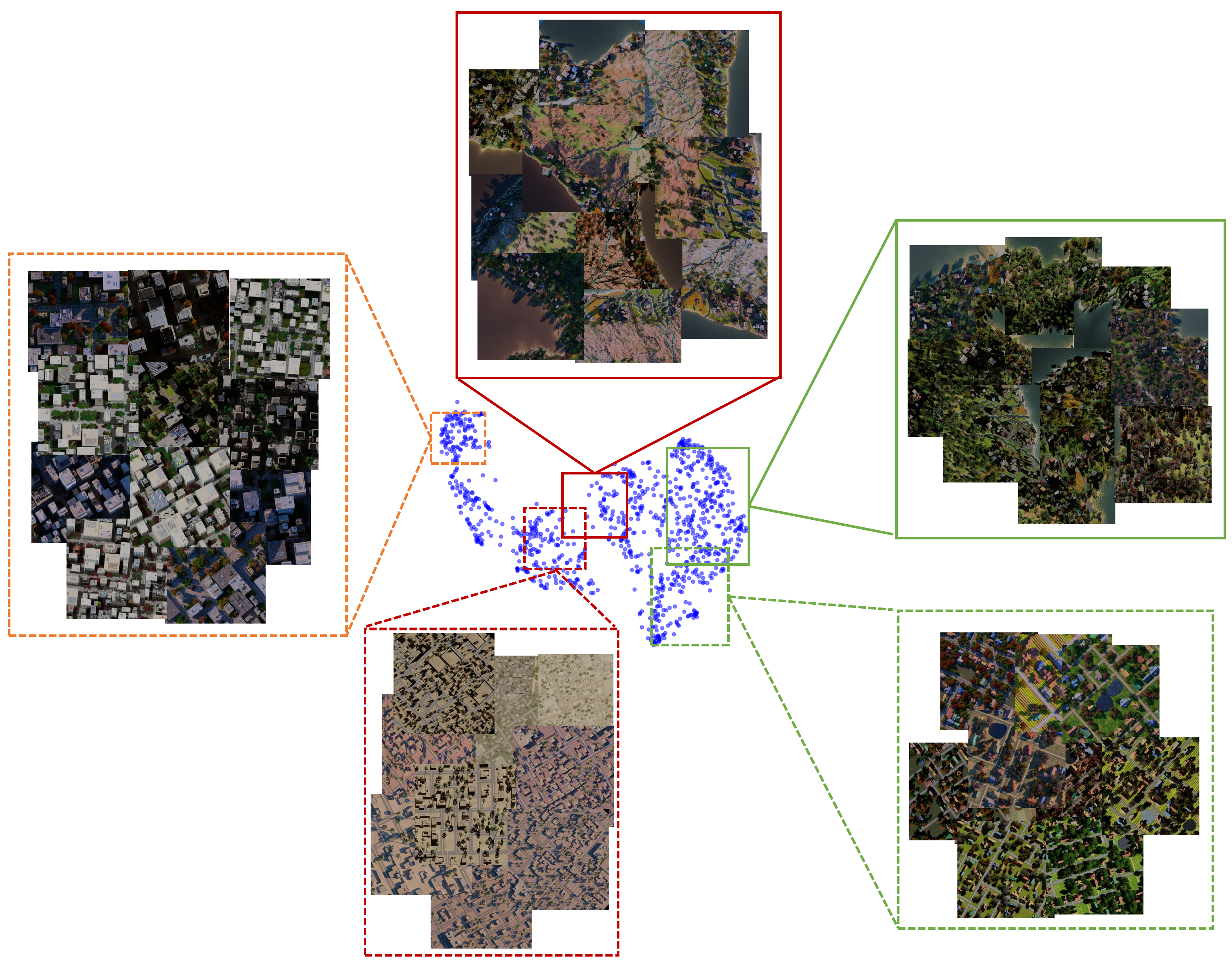}
\caption{2D UMAP of images from SyntheWorld encoded by ResNet-50. Solid lines represent images sourced from the terrain-based layout system with a GSD of 0.6-1m; dashed lines represent images sourced from the grid-based layout system with a GSD of 0.3-0.6m. \textcolor[RGB]{192,0,0}{\rule{0.3cm}{0.3cm}} represents mountainous or desert styles; \textcolor[RGB]{237,125,49}{\rule{0.3cm}{0.3cm}} represents represents urban styles; \textcolor[RGB]{112,173,71}{\rule{0.3cm}{0.3cm}} represents suburban or rural styles.}
\label{fig:umap_examples}
\end{figure*}

\cref{fig:umap_examples} presents the 2D UMAP\cite{mcinnes2018umap} visualization of the features extracted using ResNet-50~\cite{he2016deep} from the SyntheWorld dataset, along with the corresponding area image examples. During the creation of the SyntheWorld dataset, we intentionally simulated concentrated styles of regions in the real world, such as cities brimming with skyscrapers, suburbs dominated by low-rise apartments, grasslands, and farmland, as well as mountainous and desert areas characterized by earthen houses and bareland. These images of varied styles cluster together in feature space, effectively showcasing the diversity of the SyntheWorld dataset.
\section{Land Cover Texture Synthesis}
We utilized GPT-4~\cite{openai2023gpt4} along with a series of Stable Diffusion models~\cite{IF,stable2} to generate textures. Its performance far surpasses the GAN~\cite{heusel2017gans} series models in various image generation tasks. The most remarkable aspect of Stable Diffusion is its text-to-image generation model, which, given precise and detailed prompts, can produce images highly consistent with the prompt descriptions. As it was trained on the LAION-5B dataset\cite{schuhmann2022laion}, a massive dataset comprising 50 billion image-text pairs, it has ample capability to function as a texture generator, a feature already used by the dream-texture~\cite{dreamtextures} addon in the Blender community. However, if the prompts are not detailed and accurate enough, the images generated by Stable Diffusion can be highly unpredictable. 

In the early stage of SyntheWorld creation, we attempted to generate textures using simple manually written prompts, but the textures produced often lacked diversity and quality. Therefore, inspired by a YouTube video~\footnote{https://www.youtube.com/watch?v=Lu2CrEpXe0M}, we employed GPT-4 as a prompt generator for the Stable Diffusion model. 

\begin{figure*}[t!]
\centering
\includegraphics[width=\linewidth]{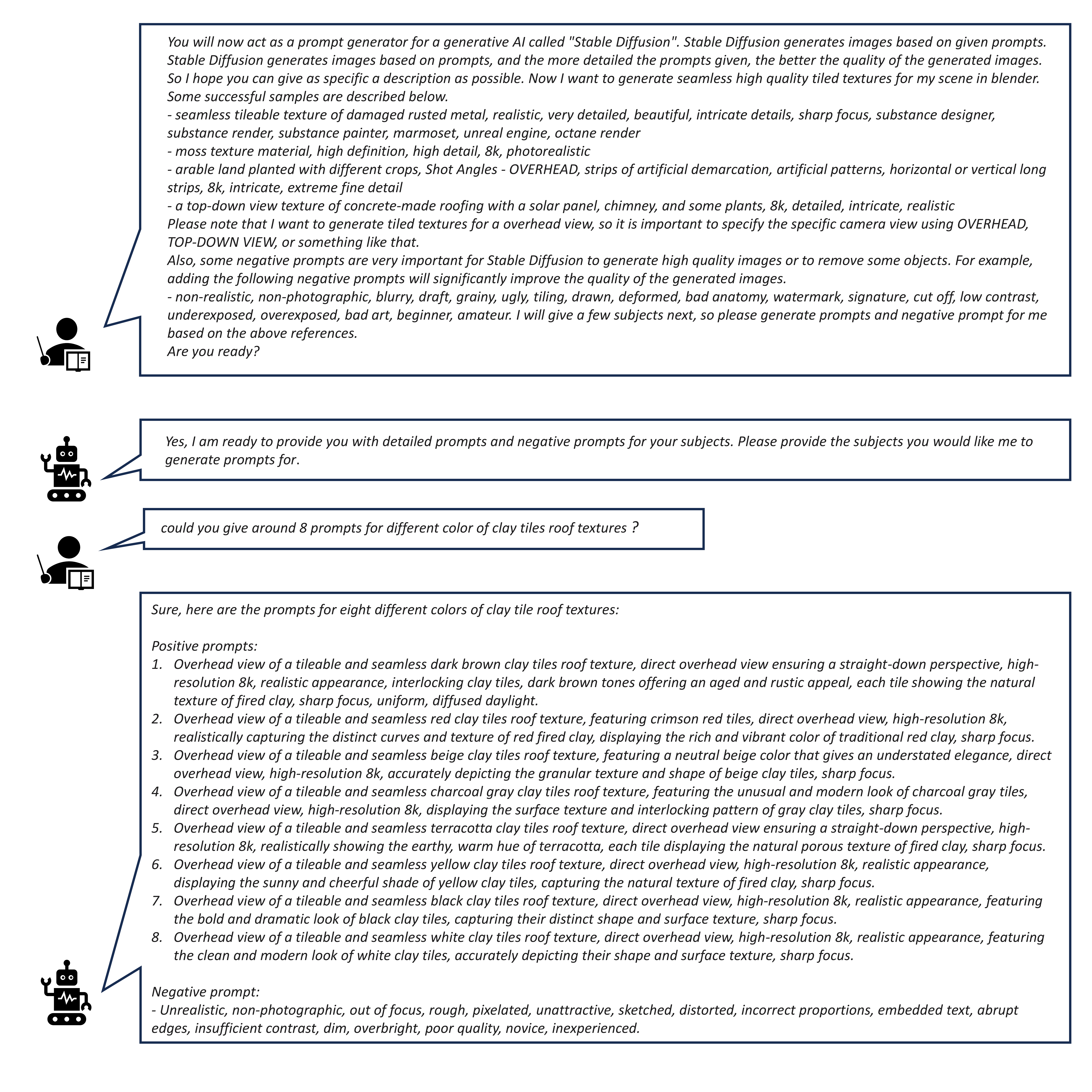}
\caption{Guide GPT-4 to generate prompts.}
\label{fig:gpt4}
\end{figure*}

As illustrated in Figure~\ref{fig:gpt4}, we initially facilitated GPT-4's understanding of the Stable Diffusion operation process by leveraging guidance from the Stable Diffusion Manual and successful prompt examples sourced from Lexia~\footnote{https://lexica.art/}. Subsequently, we would provide the subject for the texture we wish to generate, and GPT-4 would return high-quality prompts to us. This process allowed for unlimited interaction with GPT-4 to correct and refine the prompts. 

Specifically, we used the Stable Diffusion v2.1~\cite{stable2} and DeepFloyd~\cite{IF} models to generate relatively low-resolution textures, which were then upscaled to a final resolution of $2048 \times 2048$ using the Stable Diffusion x4 upscaler model~\cite{stable2}. Throughout the creation of SyntheWorld, we generated a total of 140,000 textures for seven types of geometries: roads, tree leaves, developed space, rangeland, agricultural land, bareland, and rooftops. All textures had a guidance scale of 7.5 and inference steps of 100, with each type of texture using at least 16 different GPT-4 guided prompts. 

\begin{figure*}[t!]
\centering
\includegraphics[width=\linewidth]{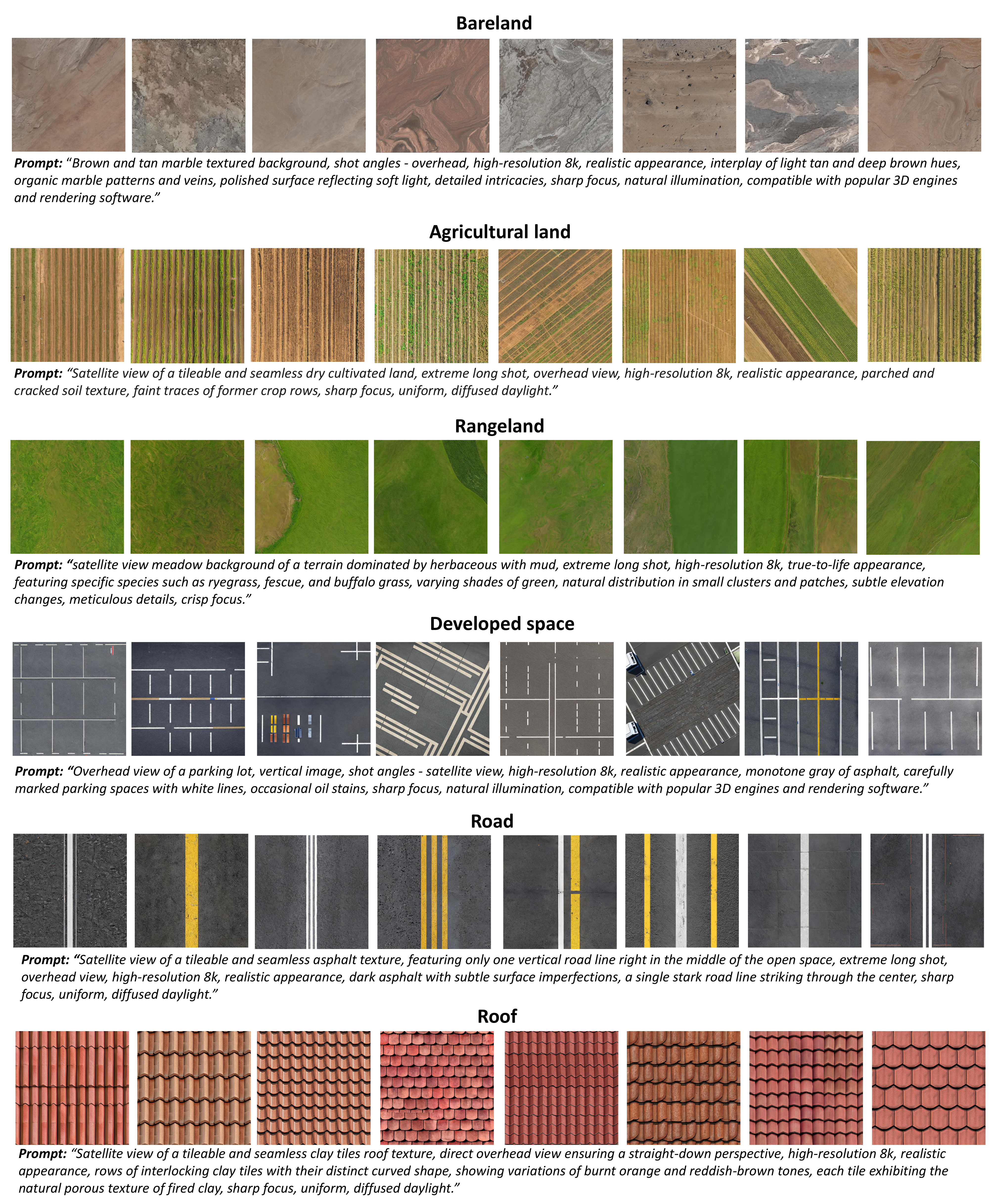}
\caption{Examples of different types of textures generated by GPT-4 guided prompts.}
\label{fig:sd_examples}
\end{figure*}

~\cref{fig:sd_examples} showcases some examples of textures generated using GPT-4 guided prompts, with all negative prompts using those provided in~\cref{fig:gpt4}. We found that the generated textures exhibited high quality in detail and rich diversity.

\section{Cross-domain Land Cover Mapping}
\subsection{Dataset Division and Experimental Settings}
~\cref{tab:division} presents the division of the dataset during our cross-domain land cover mapping experiments on the OEM~\cite{xia2023openearthmap} and LoveDA~\cite{wang2021loveda} datasets. Our experiments tested two different semantic segmentation frameworks, U-Net~\cite{ronneberger2015u} and DeepLabv3+~\cite{chen2018encoder}, along with two different CNN-based encoders, ResNet-50~\cite{he2016deep} and EfficientNet-B4~\cite{tan2019efficientnet}, as well as a transformer-based encoder~\cite{vaswani2017attention}, MiT-B5~\cite{xie2021segformer}. These models were sourced from the code repositories by Yakubovskiy~\cite{pytorchss} and Wang~\cite{lovedacode}. All experiments were carried out with a random crop of $512 \times 512$, a SGD~\cite{robbins1951stochastic} optimizer and a learning rate of 1e-3. All experiments used a batchsize of 8, in which the mixed training employed a 7:1 ratio of real to synthetic images. Each experiment was run for 100 epochs on a single Tesla A100. 
\begin{table*}[t!]
\centering
\footnotesize
\begin{tabular}{lccccccccll}
\toprule
Dataset division & \multicolumn{7}{c}{OEM} & \multicolumn{2}{c}{LoveDA} \\
\cmidrule(r){2-8} \cmidrule(lr){9-10} 
 & Africa & Asia & Central America & Europe & North America & South America & Oceana & Urban & Rural \\ 
\midrule
\# of training images & 592 & 568 & 218 & 902 & 490 & 523 & 196 & 1156 & 1366 \\ 
\# of testing images & 259 & 247 & 94 & 391 & 210 & 226 & 84 & 677 & 992 \\ 
\bottomrule
\end{tabular}
\caption{Dataset division for cross-domain land cover mapping experiments.}
\label{tab:division}
\end{table*}

\subsection{More Quantitative and Qualitative Results}
\begin{figure*}[t!]
\centering
\begin{tabular}{c}
\includegraphics[width=10.5cm]{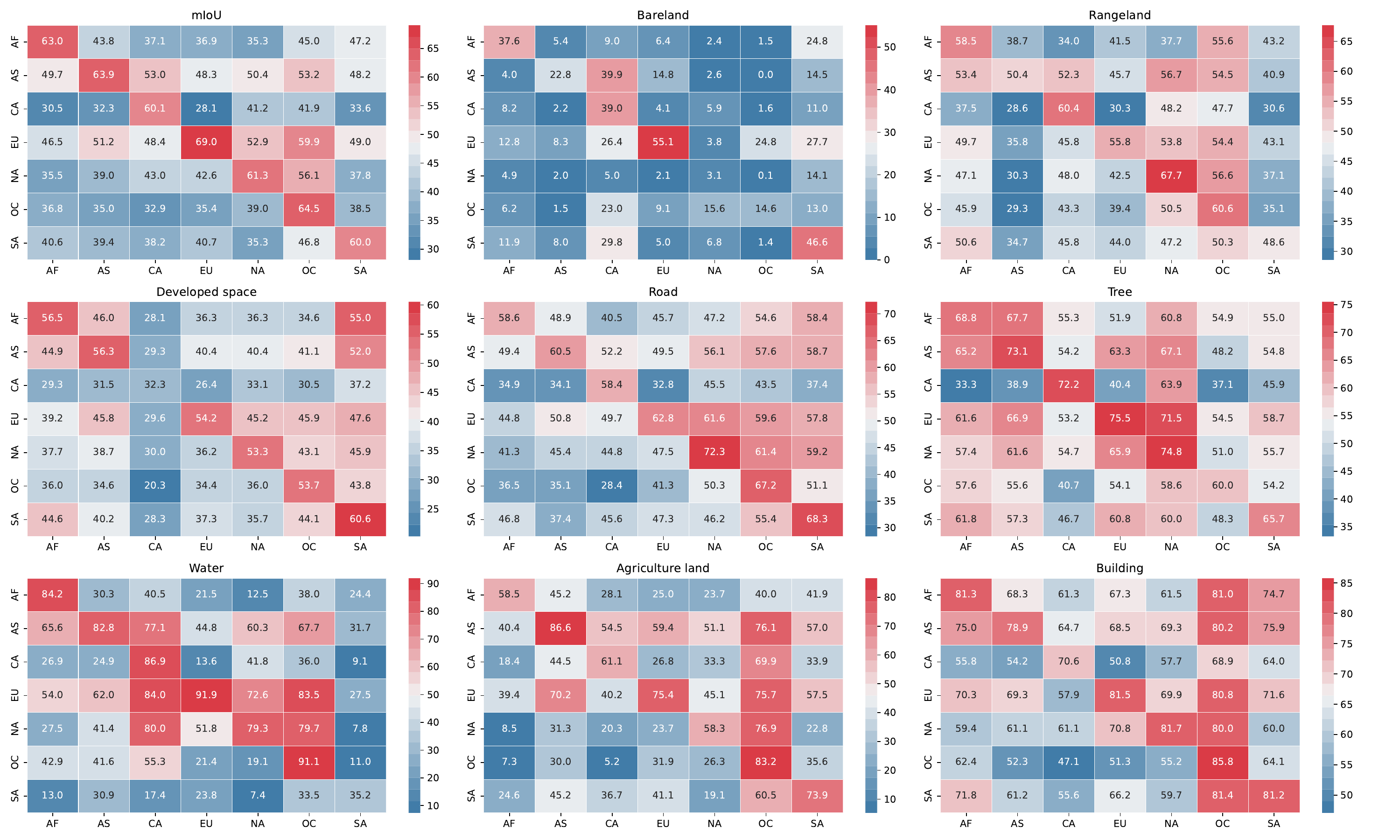}
\\(a) Without SyntheWorld.
\end{tabular}

\begin{tabular}{c}
\includegraphics[width=10.5cm]{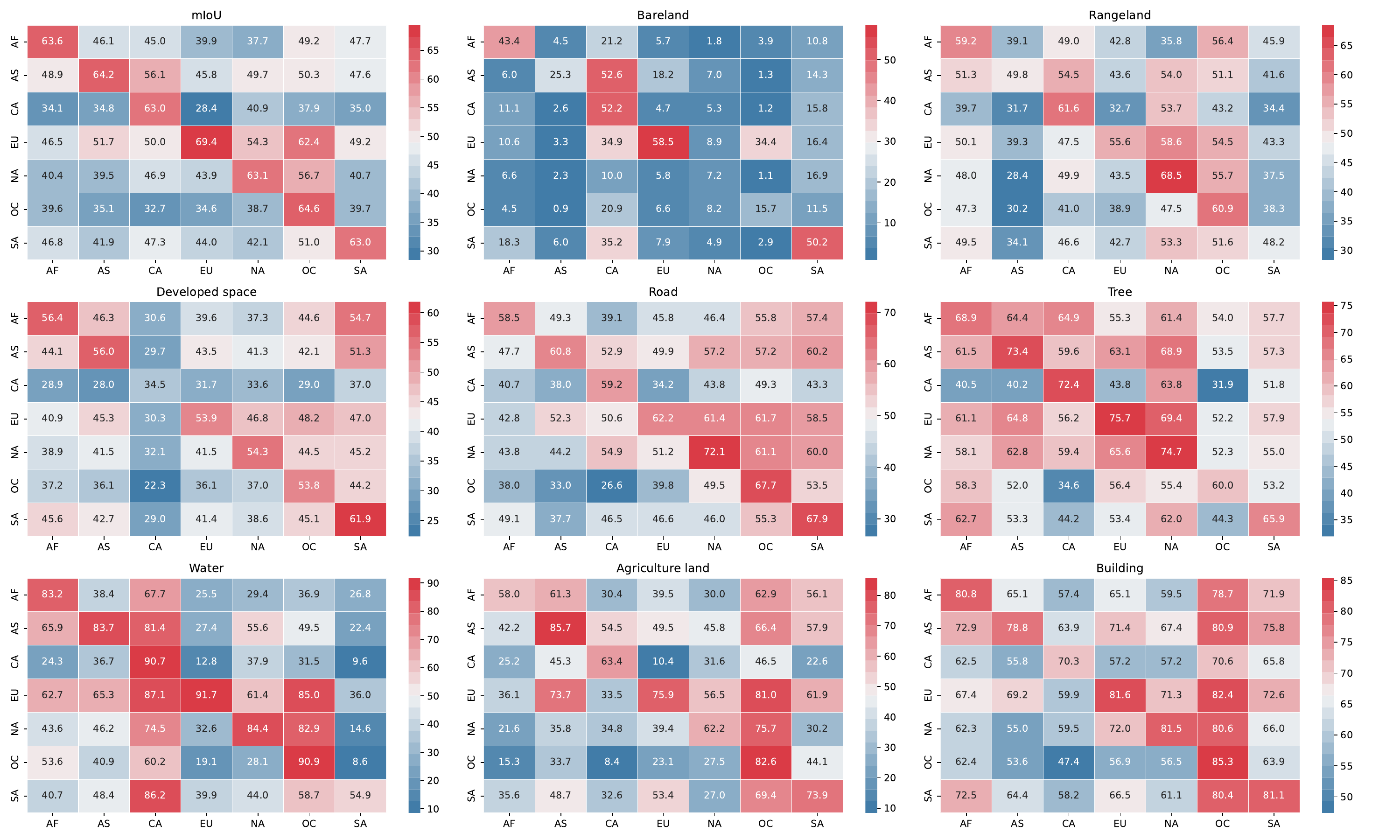}
\\(b) With SyntheWorld.
\end{tabular}

\begin{tabular}{c}
\includegraphics[width=10.5cm]{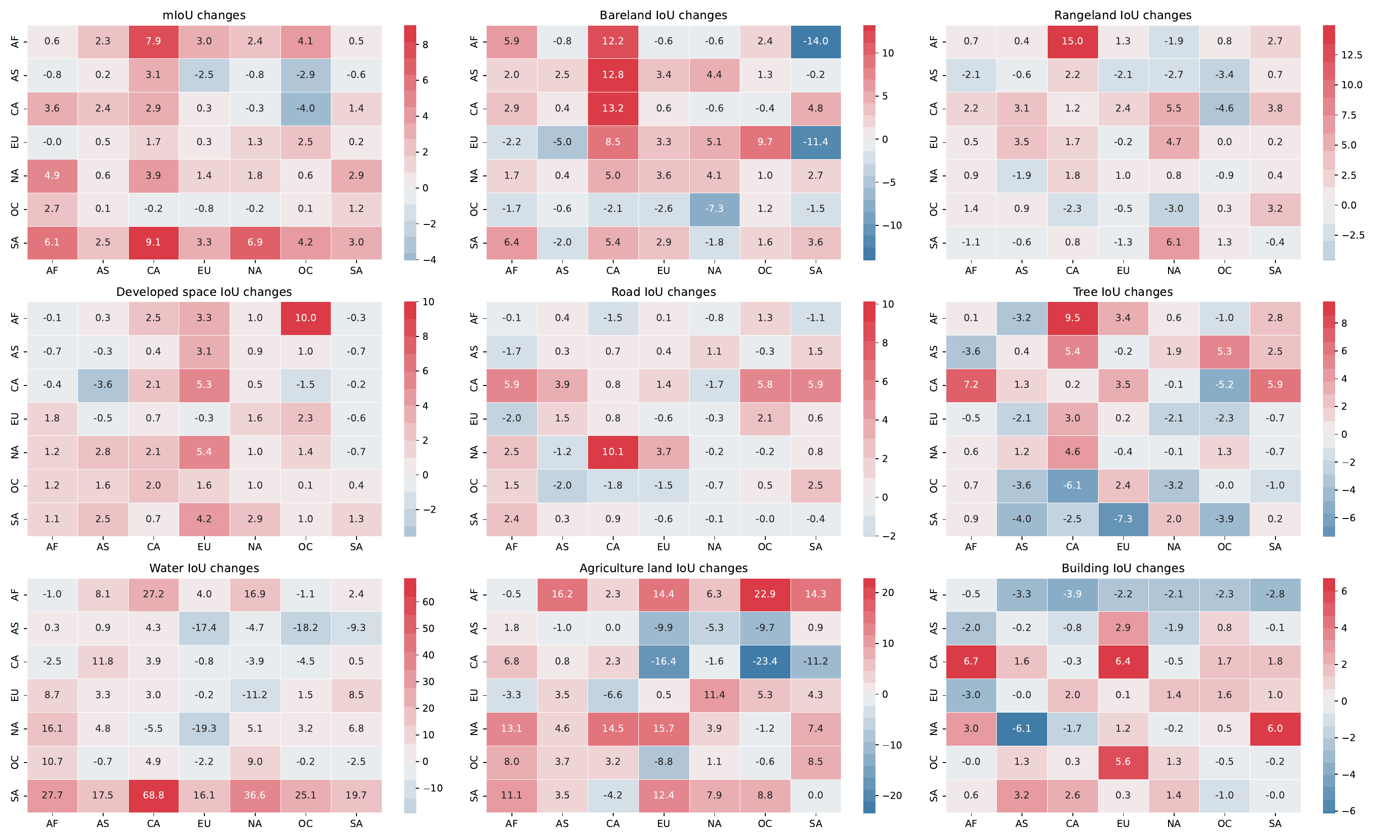}
\\(c) Performance changes.
\end{tabular}
\caption{Results of continent-wise in-domain and out-of-domain land cover mapping experiments of OEM dataset. The x-axis represents the target domain, and the y-axis represents the source domain. U-Net with EfficientNet-B4 encoder is used for all experiments.}
\label{fig:uneteffi}
\end{figure*}

~\cref{fig:uneteffi} respectively depict the mIoU and the IoU for each category, during the continent-wise experiment with the U-Net model using the EfficientNet-B4 encoder, without employing SyntheWorld; with SyntheWorld; and the changes in mIoU and IoU for each category, respectively. We use AS to represent Asia, AF for Africa, CA for Central America, EU for Europe, NA for North America, SA for South America, and OC for Oceania.

\begin{figure*}[t!]
\centering
\begin{tabular}{c}
\includegraphics[width=10.5cm]{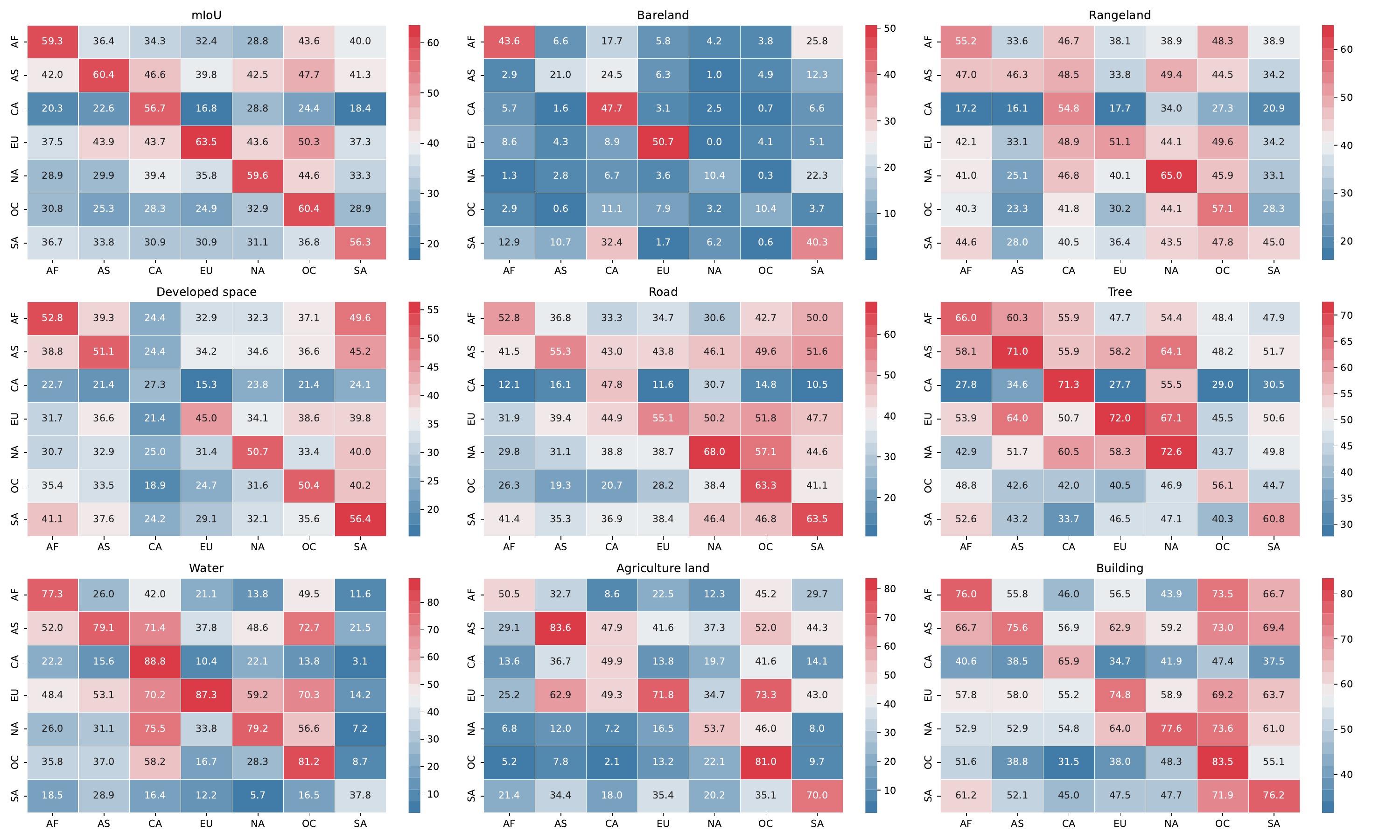}
\\(a) Without SyntheWorld.
\end{tabular}

\begin{tabular}{c}
\includegraphics[width=10.5cm]{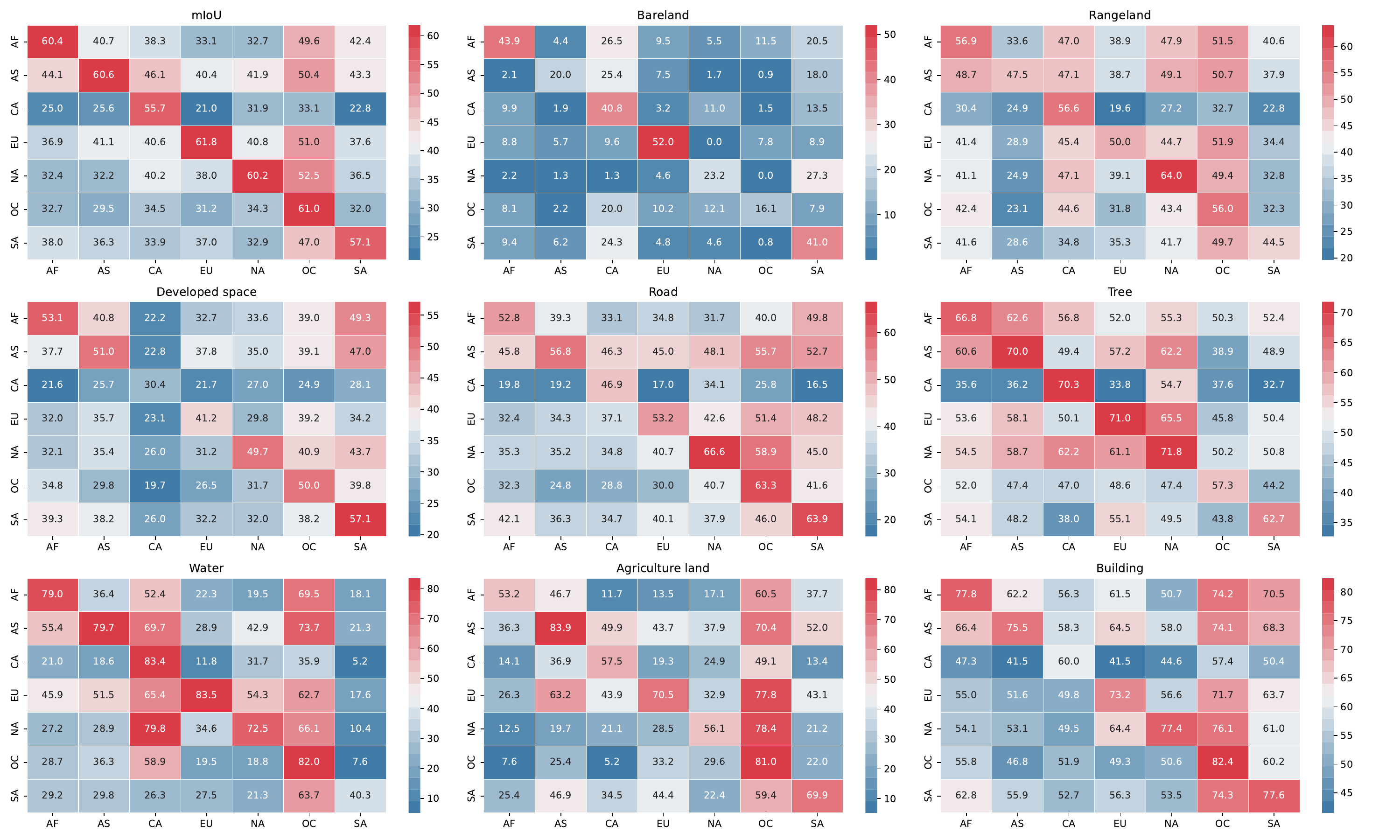}
\\(b) With SyntheWorld.
\end{tabular}

\begin{tabular}{c}
\includegraphics[width=10.5cm]{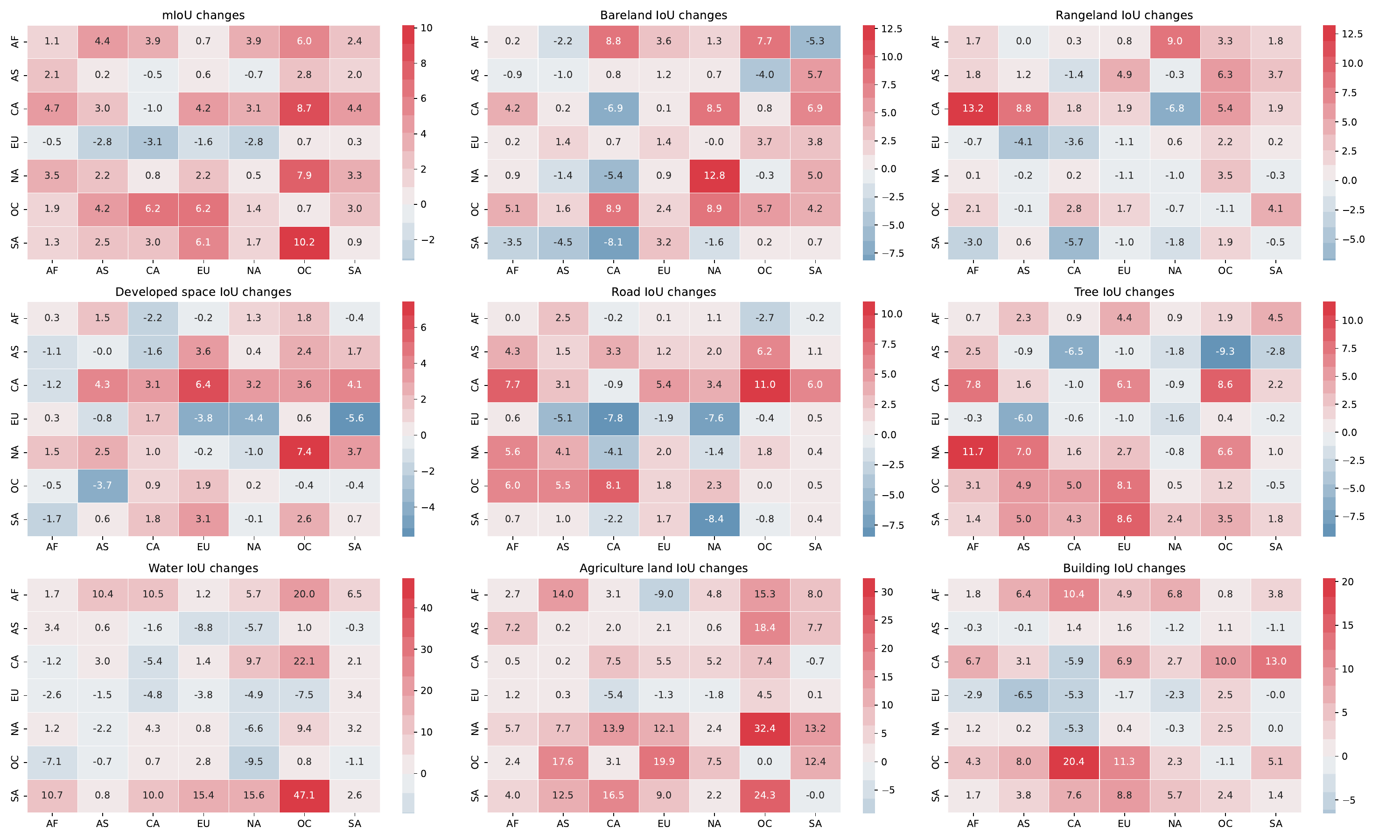}
\\(c) Performance changes.
\end{tabular}
\caption{Results of continent-wise in-domain and out-of-domain land cover mapping experiments of OEM dataset. The x-axis represents the target domain, and the y-axis represents the source domain. DeepLabv3+ with ResNet-50 encoder is used for all experiments.}
\label{fig:deeplabr50}
\end{figure*}

~\cref{fig:deeplabr50} separately present the mIoU and the IoU for each category during the continent-wise experiment using the U-Net model with the EfficientNet-B4 encoder. Specifically,~\cref{fig:deeplabr50} (a) illustrates the mIoU and IoU per category when SyntheWorld is not utilized; ~\cref{fig:deeplabr50} (b) shows the mIoU and IoU per category when SyntheWorld is employed; and~\cref{fig:deeplabr50} (c) shows the changes in both mIoU and IoU for each category.

\begin{figure*}[t!]
\centering
\begin{tabular}{c}
\includegraphics[width=10.5cm]{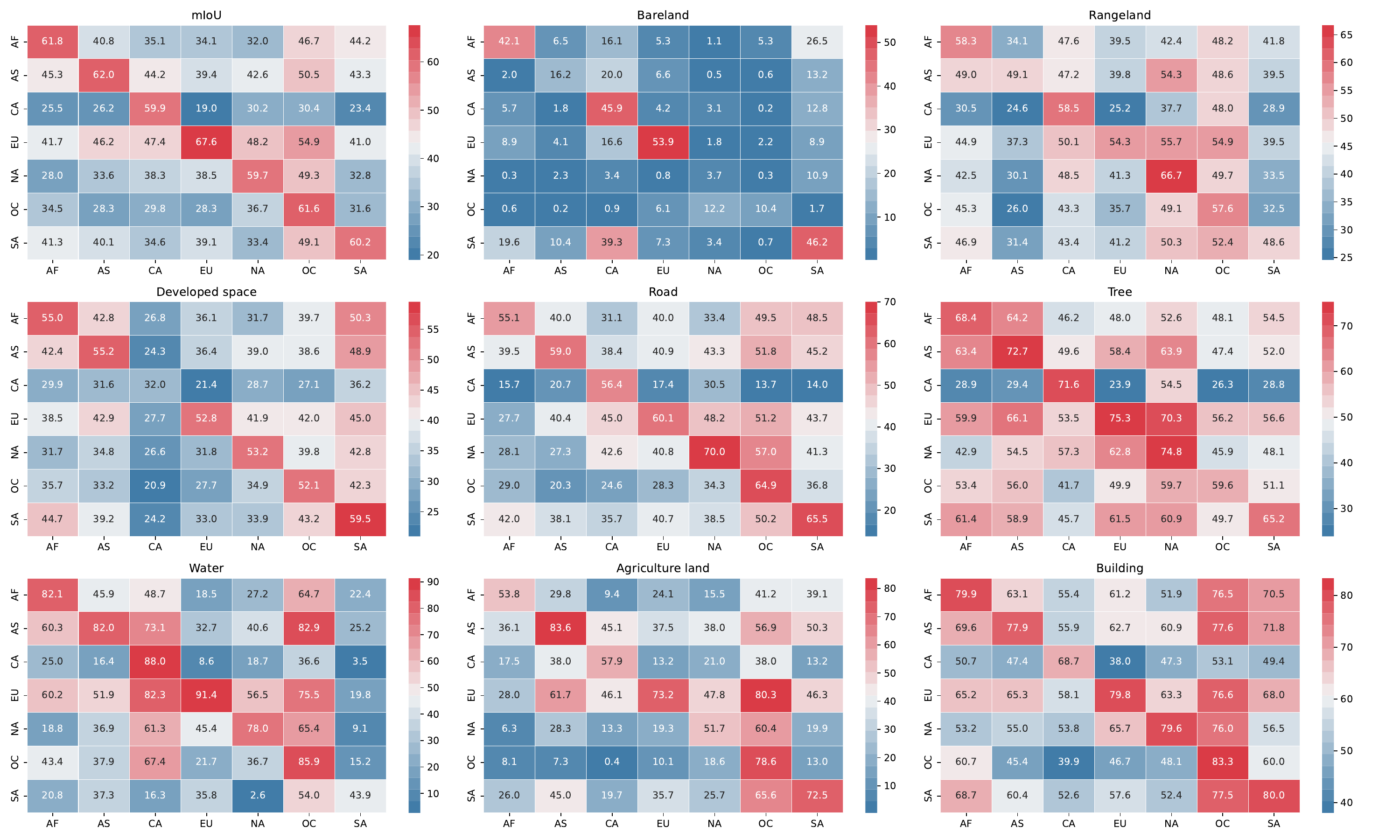}
\\(a) Without SyntheWorld.
\end{tabular}

\begin{tabular}{c}
\includegraphics[width=10.5cm]{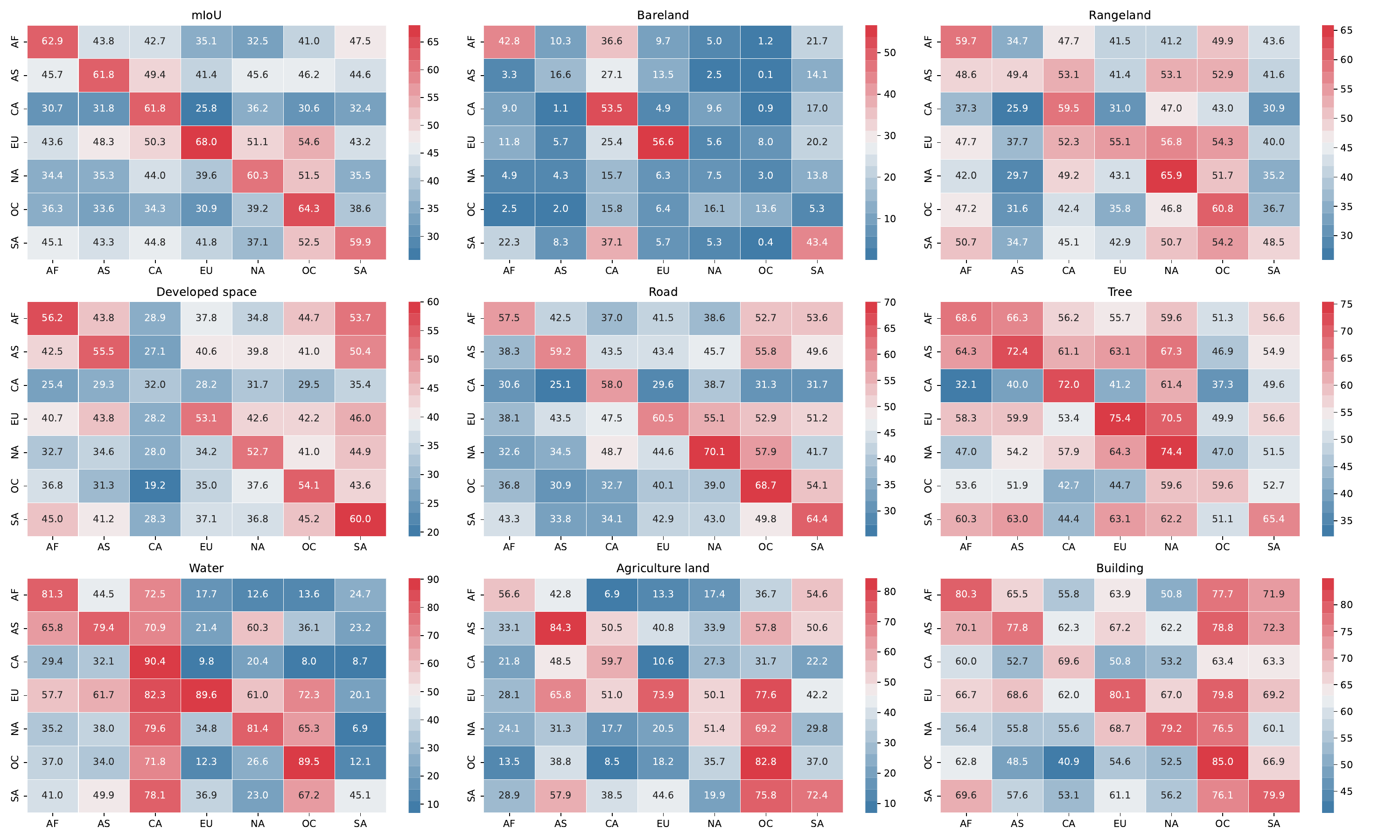}
\\(b) With SyntheWorld.
\end{tabular}

\begin{tabular}{c}
\includegraphics[width=10.5cm]{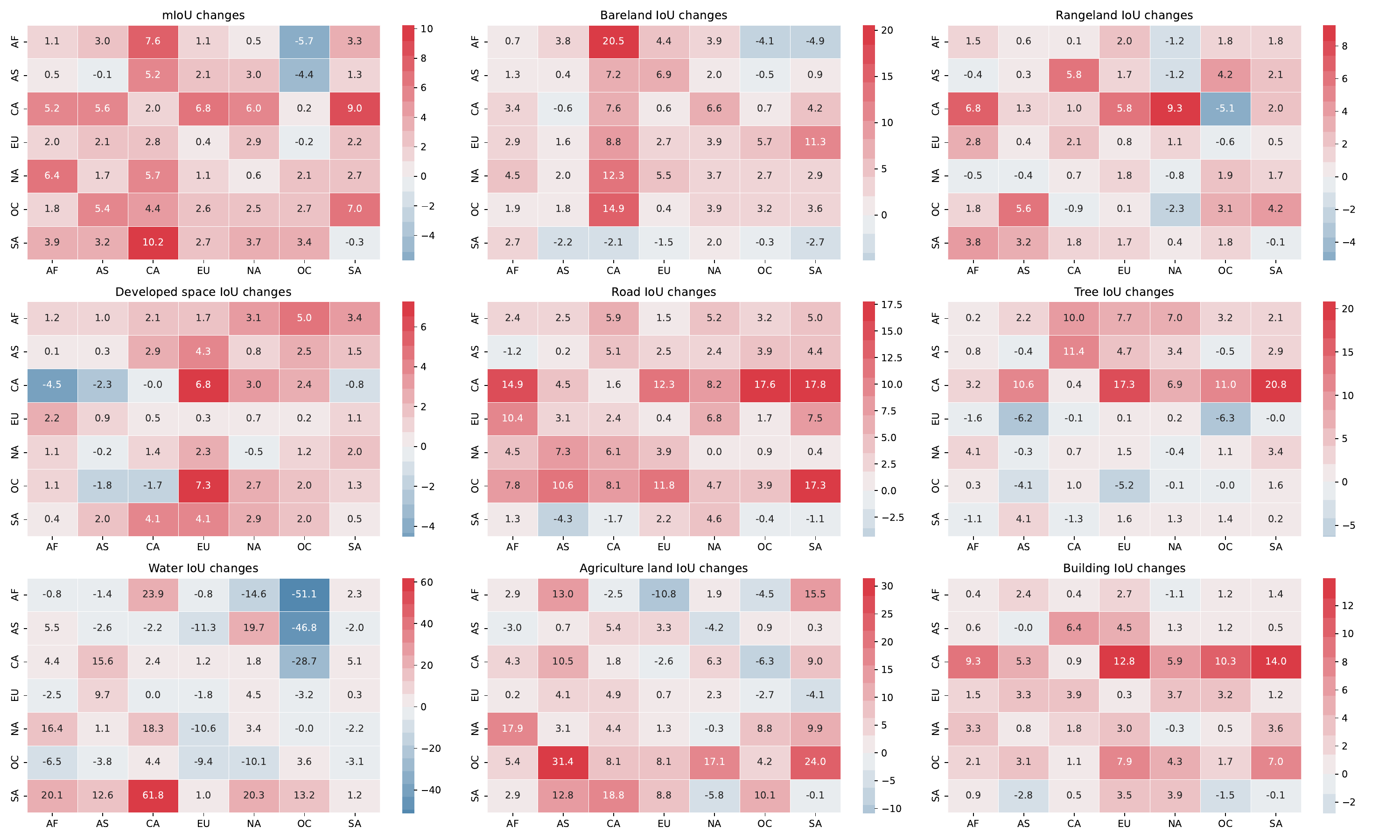}
\\(c) Performance changes.
\end{tabular}
\caption{Results of continent-wise in-domain and out-of-domain land cover mapping experiments of OEM dataset. The x-axis represents the target domain, and the y-axis represents the source domain. U-Net with MiT-B5 encoder is used for all experiments.}
\label{fig:unetmit}
\end{figure*}

~\cref{fig:unetmit} showcase the mIoU and IoU for each category when the U-Net model employing the MiT-B5 encoder is utilized in the continent-wise experiment. They depict the scenarios where SyntheWorld is not in use, where it is incorporated, and the corresponding changes in mIoU and IoU for each category, respectively.

\begin{figure*}[t!]
\centering
\begin{tabular}{cc}
\includegraphics[width=8.2cm]{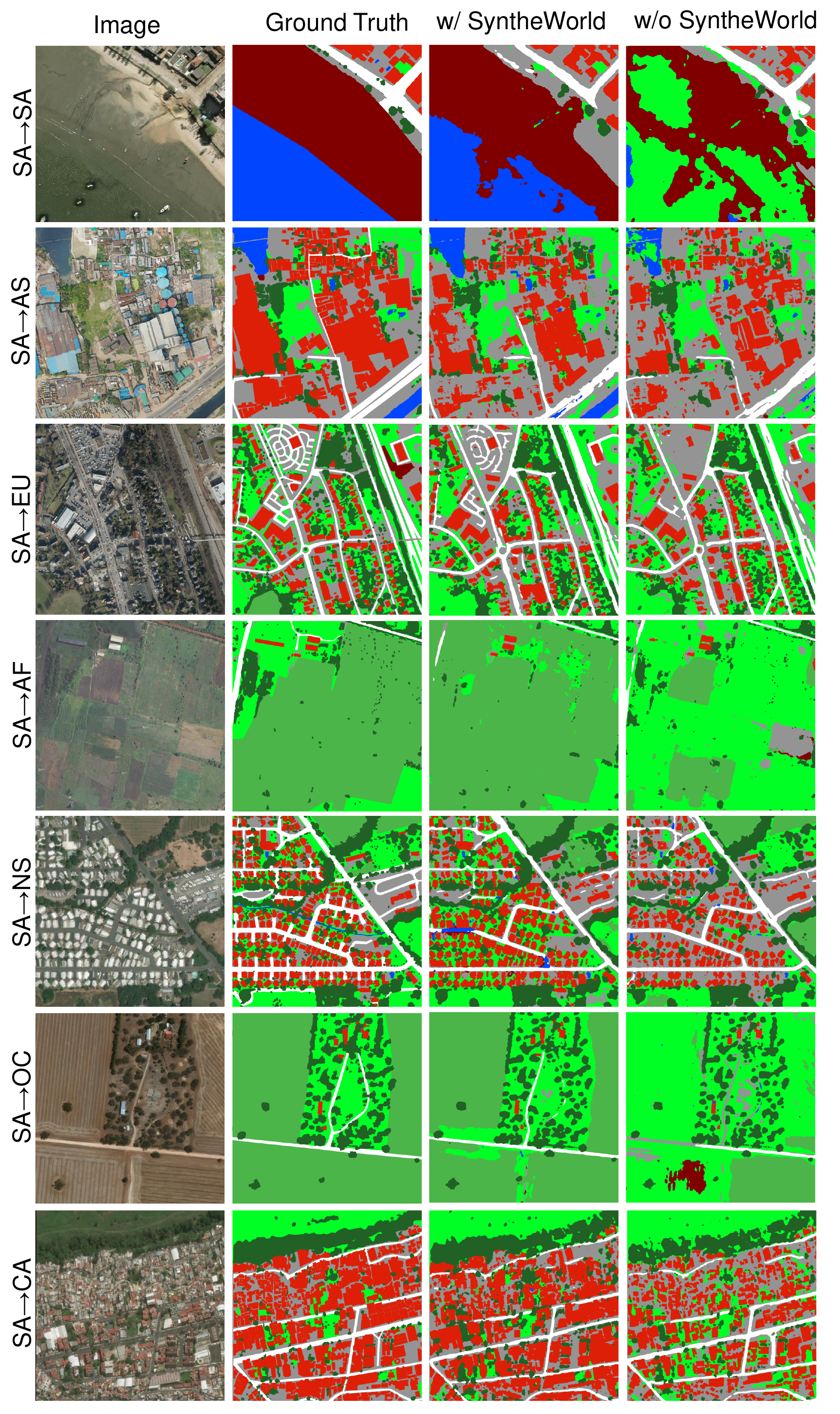}&
\includegraphics[width=7.95cm]{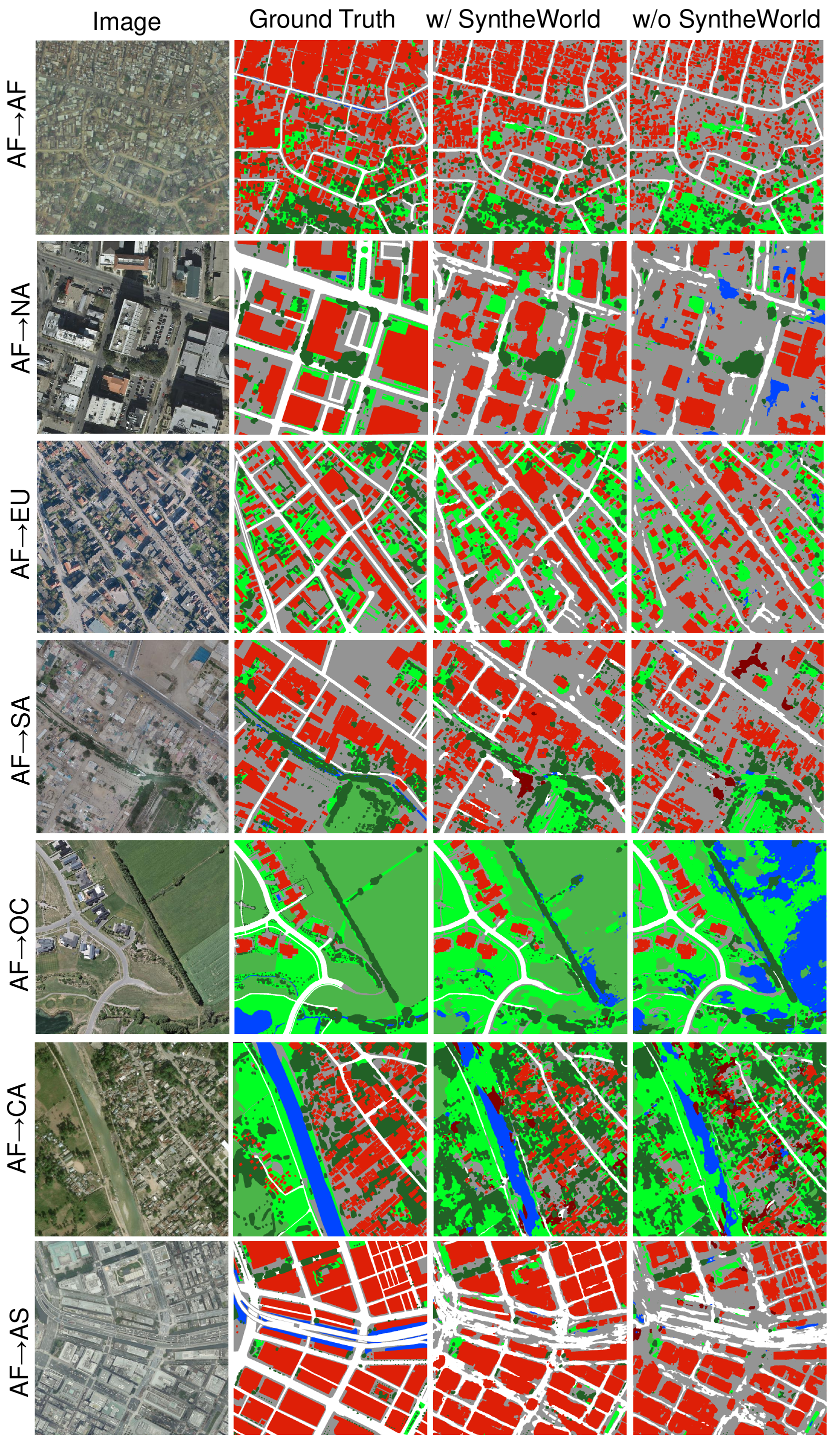}
\\(a) UNet-EfficientNet-B4&(b) DeepLabv3+ with ResNet-50
\end{tabular}
\caption{Qualitative results of continent-wise in-domain and out-of-domain land cover mapping experiments using different models on the OEM dataset.}
\label{fig:sup_vis_continent}
\end{figure*}

We can observe that for most pairs of datasets, the SyntheWorld dataset brings substantial improvement to different models. Meanwhile, ~\cref{fig:sup_vis_continent} shows more visualization results of the performance improvement brought about by SyntheWorld for different pairs of datasets through different models. This further validates the robustness of the additional knowledge brought about by the SyntheWorld dataset to real datasets.

To better interpret the experimental results in~\cref{fig:uneteffi,fig:unetmit,fig:deeplabr50}, we further analyze the average improvements in IoU for different land cover categories and the improvement of mIoU in out-of-domain testing experiments in the continent-wise setup, which are detailed in~\cref{tab:avg}.

Across different models and land cover categories, the U-Net model with the MiT-B5 backbone appears to offer the most improvement in terms of average IoU in the Bareland, Rangeland, Developed Space, Road and Tree categories. However, the U-Net model with the EfficientNet-B4 backbone has the highest IoU improvement for the 'Water' category. In terms of the category of Agriculture Land, the DeepLabv3+ with ResNet-50 backbone model shines with the highest IoU improvement. Similarly, the DeepLabv3+ with ResNet-50 model has the highest improvement in IoU in the Building category. Furthermore, considering the average improvement in mIoU, the U-Net model with MiT-B5 backbone appears to outperform the other two models.
\begin{table*}[t!]
\centering
\scriptsize
\begin{tabular}{llccccccccc}
\toprule
Methods & Backbone & \multicolumn{8}{c}{IoU(\%)} & mIoU(\%)\\
\cmidrule(lr){3-10}
& & Bareland & Rangeland & Developed space & Road & Tree & Water & Agriculture land & Building \\ 
\midrule
U-Net & EfficientNet-B4 & 1.10 & 0.96 & 1.40 & 0.97 & 0.30 & \textbf{5.81} & 3.17 & 0.51 & 1.78 \\ 
U-Net & MiT-B5 & \textbf{3.46} & \textbf{1.68} & \textbf{1.68} & \textbf{5.53} & \textbf{2.87} & 1.26 & 5.44 & 3.30 & \textbf{3.16} \\ 
DeepLabv3+ & ResNet-50 & 1.53 & 1.24 & 0.95 & 1.46 & 2.15 & 3.88 & \textbf{7.32} & \textbf{3.40} & 2.74 \\ 
\bottomrule
\end{tabular}
\caption{Average improvement in IoU for different land cover categories and average mIoU improvement in out-of-domain testing experiments under the continent-wise setup for different models.}
\label{tab:avg}
\end{table*}

\section{Additional Results for Building Change Detection}
~\cref{tab:bctableFC} presents the F1 scores of the FC-siam-Diff~\cite{daudt2018fully} model on three building change detection benchmark datasets: LEVIR-CD+~\cite{chen2020spatial}, SECOND~\cite{yang2020semantic}, and WHU-CD~\cite{ji2018fully}. The table compares the results of the model when run without (w / o) and with (w / o) SyntheWorld, which is not present in the main paper.

We follow a mixed training approach, maintaining a ratio of 7: 1 for real to synthetic images. We use the Adam~\cite{kingma2014adam} optimizer, setting the learning rate at 1e-3. Each experiment is conducted over 100 epochs and is executed on a Tesla A100 GPU.

Specifically, for the LEVIR-CD+ dataset, the model scored an F1 of 0.751 without SyntheWorld and improved to 0.766 with SyntheWorld. On the SECOND dataset, the model achieved an F1 score of 0.614 without SyntheWorld, and this score increased to 0.677 when SyntheWorld was used. Lastly, on the WHU-CD dataset, the F1 score of the FC-siam-Diff model was 0.812 without SyntheWorld and reached 0.840 with the use of SyntheWorld.

In all three cases, the use of SyntheWorld improved F1 scores, showing that it had a positive impact on the performance of the FC-siam-Diff~\cite{daudt2018fully} model when incorporating SyntheWorld.
\begin{table}[t!]
\centering  
    \begin{tabular}{lcc}
        \toprule
        \multirow{2}{*}{Datasets} & \multicolumn{2}{c}{FC-siam-Diff}  \\
        \cmidrule(r){2-3} 
        & w/o & w/ \\
        \midrule
        LEVIR-CD+~\cite{chen2020spatial} & 0.751 & \textbf{0.766} \\
        SECOND*~\cite{yang2020semantic} & 0.614 & \textbf{0.677} \\
        WHU-CD~\cite{ji2018fully} & 0.812 & \textbf{0.840} \\
        \bottomrule
    \end{tabular}
    \caption{F1 score resulting from the use or non-use of Synthe-
World across three building change detection benchmark datasets,
assessed with the FC-siam-Diff model. * means to use the part of building change label in SECOND.}
    \label{tab:bctableFC}
\end{table}

We also include three tables that detail the results of the same experiment set, i.e. training data is scarce, performed using the FC-siam-Diff~\cite{daudt2018fully}, ChangeFormer~\cite{bandara2022transformer}, and STANet~\cite{chen2020spatial} models.

~\cref{tab:bcratiotableFC} shows the performance of the FC-siam-Diff model when trained with different fractions of the real-world training set, both with and without SyntheWorld. Consistently, across all real-world datasets and at every percentage level, the incorporation of SyntheWorld boosts the model's performance.

Similar results are presented in ~\cref{tab:bcratiotableCF} and ~\cref{tab:bcratiotableSTANET}, which correspond to the ChangeFormer and STANet models, respectively. Again, in each case, the addition of SyntheWorld consistently enhances the model's performance across all datasets and at each level of real-world training data usage.

These results corroborate the main finding reported in the paper, reinforcing that the SyntheWorld dataset invariably provides a significant performance boost, particularly when the amount of real training data is limited. This beneficial effect is observed not only in the DTCDSCN~\cite{liu2020building} model, but also in the FC-siam-Diff, ChangeFormer, and STANet models.

\begin{table}[t!]
\centering
{\scriptsize 
    \begin{tabular}{lcccccc}
        \toprule
        \multirow{2}{*}{Datasets} & \multicolumn{2}{c}{1\%} & \multicolumn{2}{c}{5\%} & \multicolumn{2}{c}{10\%} \\
        \cmidrule(r){2-3} \cmidrule(lr){4-5} \cmidrule(l){6-7}
        & w/o & w/ & w/o & w/ & w/o & w/ \\
        \midrule
        LEVIR-CD+~\cite{chen2020spatial} & 0.414 & \textbf{0.558} & 0.635 & \textbf{0.658} & 0.686 & \textbf{0.759} \\
        SECOND*~\cite{yang2020semantic} & 0.381 & \textbf{0.444} & 0.545 & \textbf{0.590} & 0.507 & \textbf{0.603} \\
        WHU-CD~\cite{ji2018fully} & 0.420 & \textbf{0.519} & 0.680 & \textbf{0.717} & 0.732 & \textbf{0.738} \\
        \bottomrule
    \end{tabular}
} 
    \caption{Comparison of F1 scores from the FC-siam-Diff model trained with and without SyntheWorld, applied on three different real-world datasets at varying ratios of real image use. * means to use the part of building change label in SECOND.}
    \label{tab:bcratiotableFC}
\end{table}

\begin{table}[t!]
\centering
{\scriptsize 
    \begin{tabular}{lcccccc}
        \toprule
        \multirow{2}{*}{Datasets} & \multicolumn{2}{c}{1\%} & \multicolumn{2}{c}{5\%} & \multicolumn{2}{c}{10\%} \\
        \cmidrule(r){2-3} \cmidrule(lr){4-5} \cmidrule(l){6-7}
        & w/o & w/ & w/o & w/ & w/o & w/ \\
        \midrule
        LEVIR-CD+~\cite{chen2020spatial} & 0.357 & \textbf{0.469} & 0.520 & \textbf{0.581} & 0.602 & \textbf{0.671} \\
        SECOND*~\cite{yang2020semantic} & 0.329 & \textbf{0.428} & 0.483 & \textbf{0.521} & 0.503 & \textbf{0.564} \\
        WHU-CD~\cite{ji2018fully} & 0.227 & \textbf{0.290} & 0.517 & \textbf{0.644} & 0.565 & \textbf{0.665} \\
        \bottomrule
    \end{tabular}
} 
    \caption{Comparison of F1 scores from the ChangeFormer model trained with and without SyntheWorld, applied on three different real-world datasets at varying ratios of real image use. * means to use the part of building change label in SECOND.}
    \label{tab:bcratiotableCF}
\end{table}

\begin{table}[t!]
\centering
{\scriptsize   
    \begin{tabular}{lcccccc}
        \toprule
        \multirow{2}{*}{Datasets} & \multicolumn{2}{c}{1\%} & \multicolumn{2}{c}{5\%} & \multicolumn{2}{c}{10\%} \\
        \cmidrule(r){2-3} \cmidrule(lr){4-5} \cmidrule(l){6-7}
        & w/o & w/ & w/o & w/ & w/o & w/ \\
        \midrule
        LEVIR-CD+~\cite{chen2020spatial} & 0.541 & \textbf{0.600} & 0.575 & \textbf{0.643} & 0.688 & \textbf{0.741} \\
        SECOND*~\cite{yang2020semantic} & 0.536 & \textbf{0.546} & 0.554 & \textbf{0.628} & 0.619 & \textbf{0.645} \\
        WHU-CD~\cite{ji2018fully} & 0.295 & \textbf{0.344} & 0.524 & \textbf{0.595} & 0.687 & \textbf{0.719} \\
        \bottomrule
    \end{tabular}
}    
    \caption{Comparison of F1 scores from the STANet model trained with and without SyntheWorld, applied on three different real-world datasets at varying ratios of real image use. * means to use the part of building change label in SECOND.}
    \label{tab:bcratiotableSTANET}
\end{table}

As demonstrated in Tables~\ref{tab:BSET_LEVIR},~\ref{tab:BSET_SECOND}, and~\ref{tab:BSET_WHUCD}, we have conducted comprehensive experiments with the DTCDSCN and FC-siam-Diff models, training them on two synthetic datasets, AICD~\cite{bourdis2011constrained} and SyntheWorld, and subsequently testing these models on three real-world datasets.

In the training phase with the synthetic datasets, we utilized the Adam optimizer and set the learning rate to 5e-4. A notable observation is that synthetic data-trained models typically achieved their best performance within fewer than 50 epochs.

In order to draw a meaningful comparison, Oracle experiments were also performed, in which the models were trained directly on real-world datasets. In these instances, we adopted a higher learning rate of 1e-3 and found that the models reached their optimal performance approximately around the 100 epoch mark.

The experiments underscored a clear trend: regardless of the real-world dataset used for testing, the models trained on the SyntheWorld dataset consistently outperformed the ones trained on the AICD dataset. Moreover, the performance of SyntheWorld-trained models, while not matching the models trained directly on the real-world datasets, came close enough to indicate a significant value of the synthetic dataset in training effective change detection models. 

\begin{figure*}[t!]
\centering
\begin{tabular}{cc}
\includegraphics[width=7.5cm]{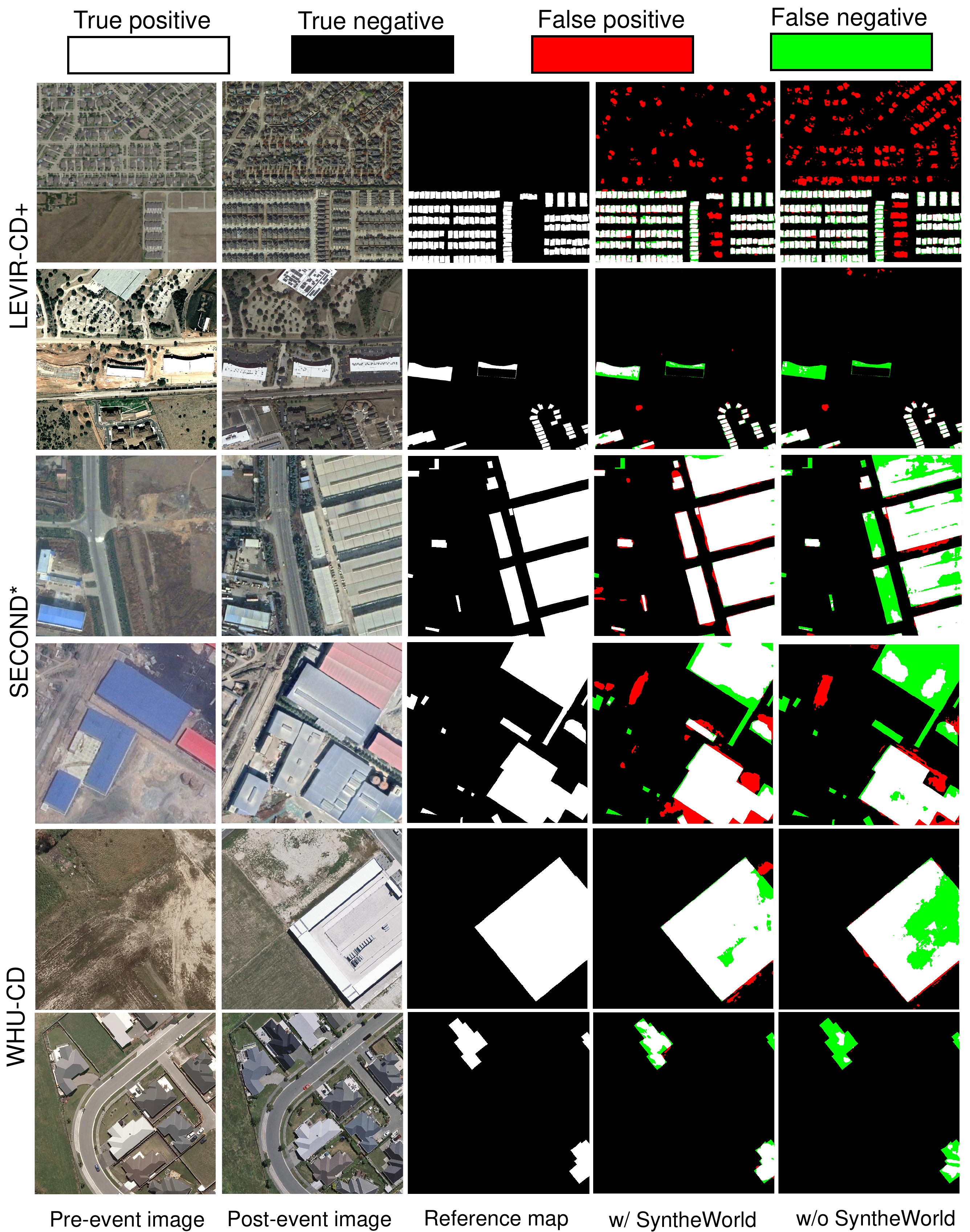}&
\includegraphics[width=7.5cm]{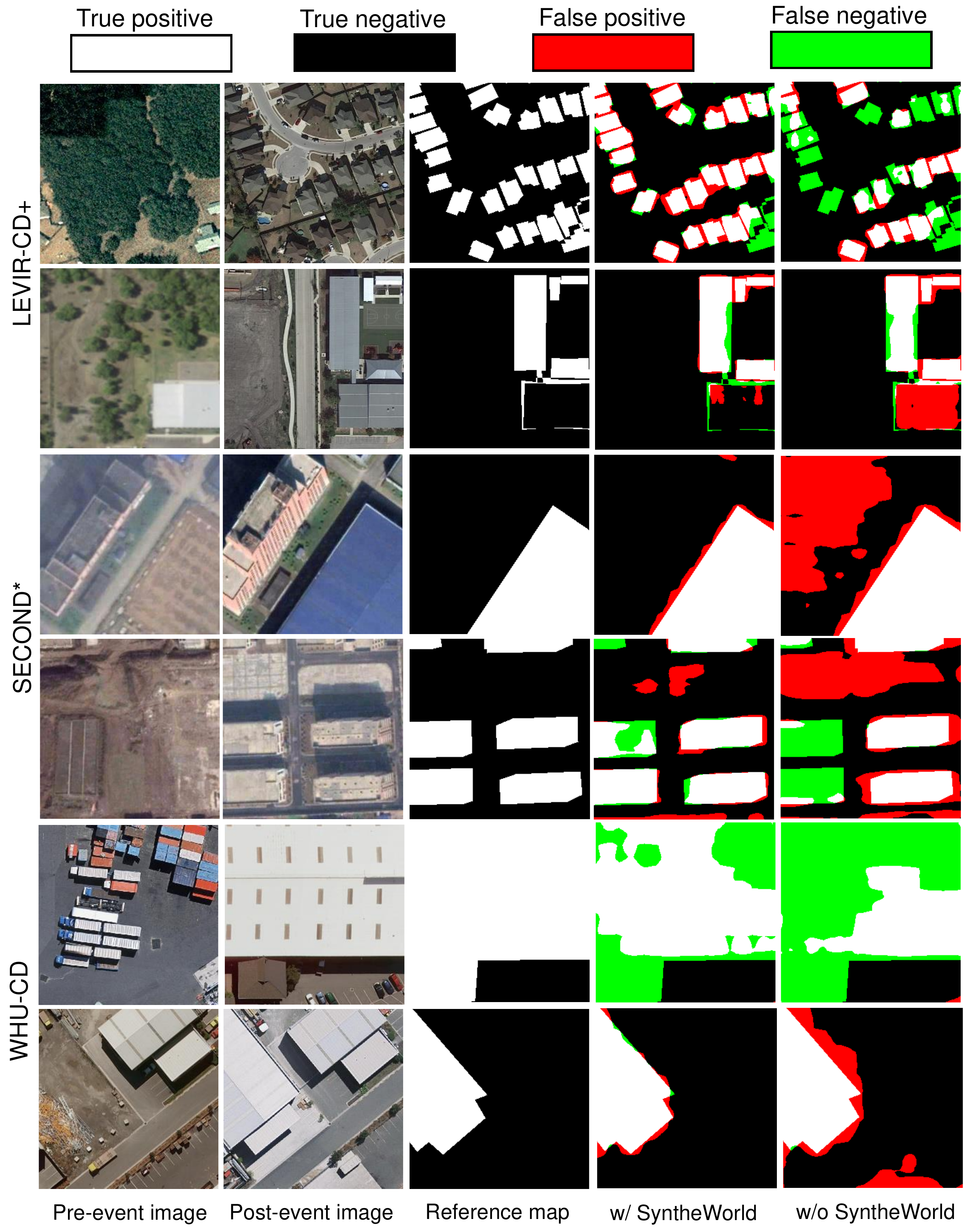}
\\(a) FC-siam-Diff&(b) STANet
\end{tabular}

\begin{tabular}{cc}
\includegraphics[width=7.5cm]{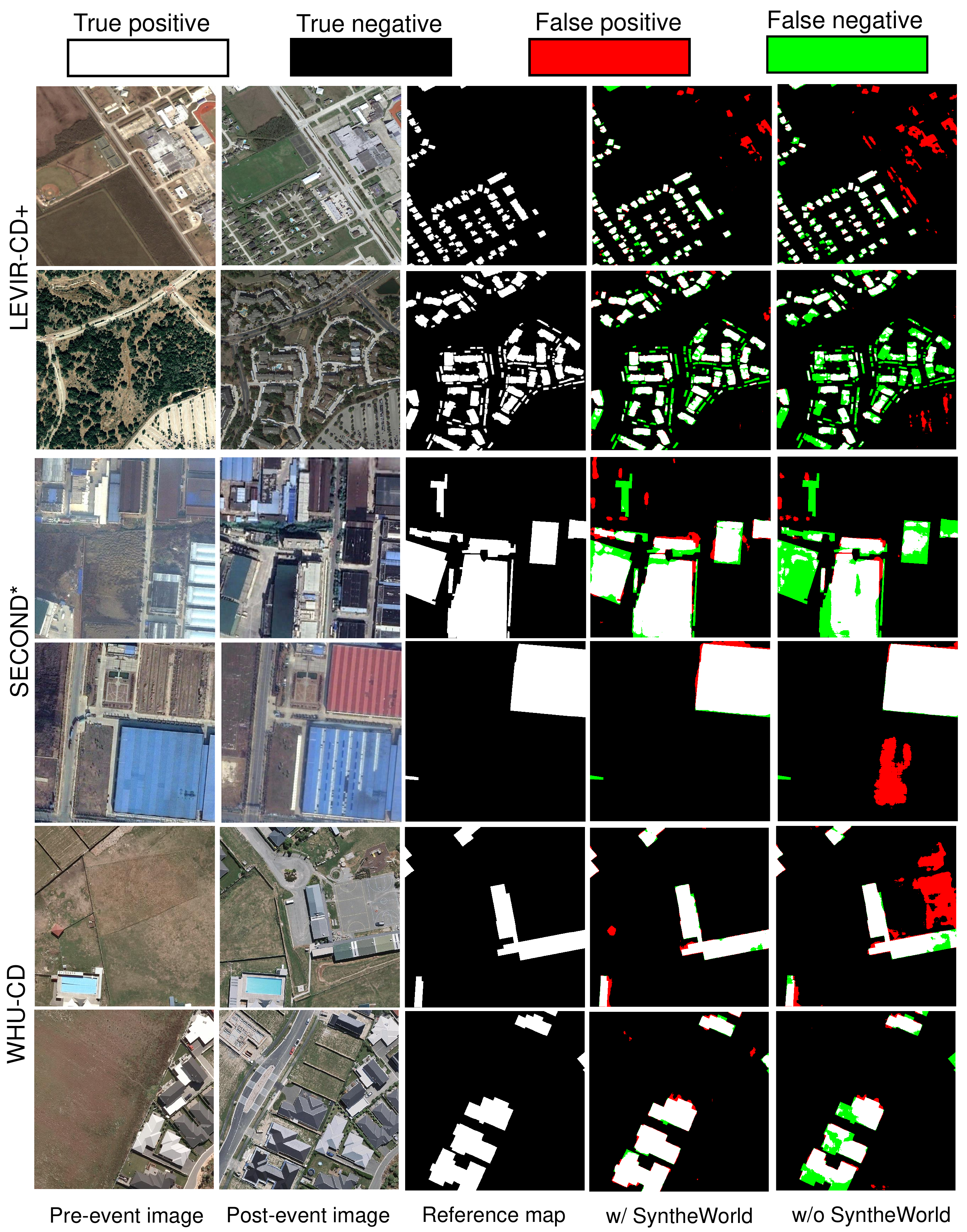}&
\includegraphics[width=7.5cm]{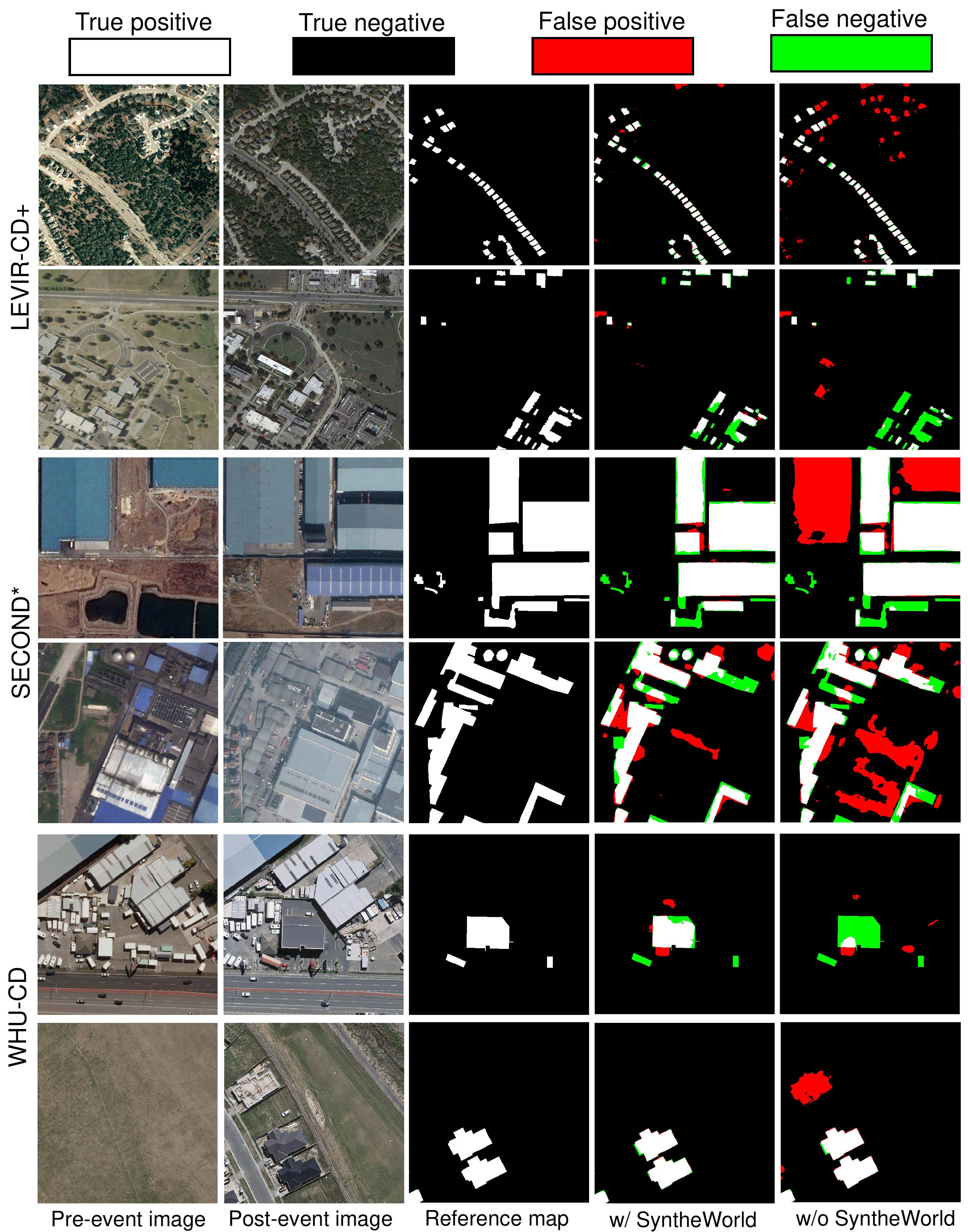}
\\(c) DTCDSCN&(d) ChangeFormer
\end{tabular}
\caption{Qualitative results of the building change detection task using different models and datasets, with and without the incorporation of the SyntheWorld dataset.}
\label{fig:sup_vis_cd}
\end{figure*}

Further support for the effectiveness of the SyntheWorld dataset can be found in~\cref{fig:sup_vis_cd}, which provides a visual comparison of model performance when trained with different datasets using different models.

This highlights the immense potential of the SyntheWorld dataset for building change detection. The ability of SyntheWorld to close the gap between synthetic and real-world data to an acceptable margin is an encouraging sign. It signifies that we could significantly reduce our reliance on large volumes of real-world data for training effective change detection models.

\begin{table}[t!]
\centering
{\footnotesize    
    \begin{tabular}{lcc}
        \toprule
        Training Data & \multicolumn{2}{c}{Models} \\
        \cmidrule(r){2-3}
        & DTCDSCN & FC-siam-Diff \\
        \midrule
        AICD~\cite{bourdis2011constrained} & 0.160 & 0.133 \\
        SyntheWorld & \textbf{0.364} & \textbf{0.425}  \\
        \midrule
        Oracle & 0.793 & 0.751  \\
        \bottomrule
    \end{tabular}
}    
    \caption{Comparison of the best F1 scores achieved on the LEVIR-CD+ dataset test set by the DTCDSCN and FC-siam-Diff models. Each model was trained on each synthetic dataset and tested on the LEVIR-CD+ dataset. The Oracle results indicate the performance of each model when trained and tested on the LEVIR-CD+ dataset.}
    \label{tab:BSET_LEVIR}
\end{table}

\begin{table}[t!]
\centering
{\footnotesize    
    \begin{tabular}{lcc}
        \toprule
        Training Data & \multicolumn{2}{c}{Models} \\
        \cmidrule(r){2-3}
        & DTCDSCN & FC-siam-Diff \\
        \midrule
        AICD~\cite{bourdis2011constrained} & 0.348 & 0.324 \\
        SyntheWorld & \textbf{0.451} & \textbf{0.461}  \\
        \midrule
        Oracle & 0.712 & 0.614  \\
        \bottomrule
    \end{tabular}
}    
    \caption{Comparison of the best F1 scores obtained on the SECOND dataset test set by the DTCDSCN and FC-siam-Diff models. The models were independently trained on each synthetic dataset and subsequently tested on the SECOND dataset. The Oracle performance is derived from training and testing each model on the SECOND dataset.}
    \label{tab:BSET_SECOND}
\end{table}

\begin{table}[t!]
\centering
{\footnotesize    
    \begin{tabular}{lcc}
        \toprule
        Training Data & \multicolumn{2}{c}{Models} \\
        \cmidrule(r){2-3}
        & DTCDSCN & FC-siam-Diff \\
        \midrule
        AICD~\cite{bourdis2011constrained} & 0.231 & 0.236 \\
        SyntheWorld & \textbf{0.550} & \textbf{0.540}  \\
        \midrule
        Oracle & 0.769 & 0.812  \\
        \bottomrule
    \end{tabular}
}    
    \caption{Performance comparison of the best F1 scores on the WHU-CD test set, attained by the DTCDSCN and FC-siam-Diff models. The Oracle results represent the performance of the models when both training and testing are performed on the WHU-CD dataset.}
    \label{tab:BSET_WHUCD}
\end{table}

\section{Attribution of Utilized Blender Addons}
The development process of SyntheWorld is based on Blender 3.4. We utilized and modified a multitude of community addons, combining them for the generation of SyntheWorld. Detailed information on all the addons used during our development process is shown in~\cref{tab:addons}.

\begin{table*}[t!]
\centering
\tiny
\begin{tabular}{lllll}
\toprule
\textbf{Name} & \textbf{Author} & \textbf{Version} & \textbf{License} & \textbf{URL} \\
\midrule
Realtime River Generator & specoolar & 1.1 & RF & \url{https://blendermarket.com/products/river-generator} \\
Next Street & Next Realm & 2.0 & RF & \url{https://blendermarket.com/products/next-street} \\
Objects Replacer & Georeality Design & 1.06 & GPL & \url{https://blendermarket.com/products/objects-replacer/docs} \\
Albero & Greenbaburu & 0.3 & RF & \url{https://blendermarket.com/products/albero---geometry-nodes-powered-tree-generator} \\
Hira Building Generator & HiranojiStore & 0.9 & RF & \url{https://blendermarket.com/products/hira-building-generator} \\
Procedural Building Generator & Isak Waltin & 1.2.1 & CC-BY 4.0 & \url{https://blendermarket.com/products/building-gen} \\
Pro Atmo & Contrastrender & 1.0 & GPL & \url{https://blendermarket.com/products/pro-atmo} \\
Modular Buildings Creator & PH Felix & 1.0 & RF & \url{https://blendermarket.com/products/modular-buildings-creator} \\
Next Trees & Next Realm & 2.0 & RF & \url{https://blendermarket.com/products/next-trees} \\
SceneCity & Arnaud & 1.9.3 & RF & \url{http://www.cgchan.com/store/scenecity} \\
Flex Road Generator & EasyNodes & 1.1.0 & RF & \url{https://www.cgtrader.com/3d-models/scripts-plugins/modelling/blender-mesh-curve-to-road} \\
Buildify & Pavel Oliva & 1.0 & RF & \url{https://paveloliva.gumroad.com/l/buildify} \\

\bottomrule
\end{tabular}
\caption{List of Blender Addons used in the study.}
\label{tab:addons}
\end{table*}

%% file: arxiv.bbl
\begin{thebibliography}{10}\itemsep=-1pt

\bibitem{alvarez2020geometric}
David Alvarez-Melis and Nicolo Fusi.
\newblock Geometric dataset distances via optimal transport.
\newblock {\em Advances in Neural Information Processing Systems},
  33:21428--21439, 2020.

\bibitem{bandara2022transformer}
Wele Gedara~Chaminda Bandara and Vishal~M Patel.
\newblock A transformer-based siamese network for change detection.
\newblock In {\em IGARSS 2022-2022 IEEE International Geoscience and Remote
  Sensing Symposium}, pages 207--210. IEEE, 2022.

\bibitem{bourdis2011constrained}
Nicolas Bourdis, Denis Marraud, and Hichem Sahbi.
\newblock Constrained optical flow for aerial image change detection.
\newblock In {\em 2011 IEEE international geoscience and remote sensing
  symposium}, pages 4176--4179. IEEE, 2011.

\bibitem{butler2012naturalistic}
Daniel~J Butler, Jonas Wulff, Garrett~B Stanley, and Michael~J Black.
\newblock A naturalistic open source movie for optical flow evaluation.
\newblock In {\em Computer Vision--ECCV 2012: 12th European Conference on
  Computer Vision, Florence, Italy, October 7-13, 2012, Proceedings, Part VI
  12}, pages 611--625. Springer, 2012.

\bibitem{chen2020spatial}
Hao Chen and Zhenwei Shi.
\newblock A spatial-temporal attention-based method and a new dataset for
  remote sensing image change detection.
\newblock {\em Remote Sensing}, 12(10):1662, 2020.

\bibitem{chen2018encoder}
Liang-Chieh Chen, Yukun Zhu, George Papandreou, Florian Schroff, and Hartwig
  Adam.
\newblock Encoder-decoder with atrous separable convolution for semantic image
  segmentation.
\newblock In {\em Proceedings of the European conference on computer vision
  (ECCV)}, pages 801--818, 2018.

\bibitem{blender}
Blender~Online Community.
\newblock {\em Blender - a 3D modelling and rendering package}.
\newblock Blender Foundation, Stichting Blender Foundation, Amsterdam, 2018.

\bibitem{ProAtmo}
Contrastrender.
\newblock Pro atmo: User documentation.
\newblock Blender Addons, 2022.
\newblock Accessed: 2023-03-24.

\bibitem{Cordts2016Cityscapes}
Marius Cordts, Mohamed Omran, Sebastian Ramos, Timo Rehfeld, Markus Enzweiler,
  Rodrigo Benenson, Uwe Franke, Stefan Roth, and Bernt Schiele.
\newblock The cityscapes dataset for semantic urban scene understanding.
\newblock In {\em Proc. of the IEEE Conference on Computer Vision and Pattern
  Recognition (CVPR)}, 2016.

\bibitem{daudt2018fully}
Rodrigo~Caye Daudt, Bertrand Le~Saux, and Alexandre Boulch.
\newblock Fully convolutional siamese networks for change detection.
\newblock In {\em 2018 25th IEEE International Conference on Image Processing
  (ICIP)}, pages 4063--4067. IEEE, 2018.

\bibitem{demir2018deepglobe}
Ilke Demir, Krzysztof Koperski, David Lindenbaum, Guan Pang, Jing Huang, Saikat
  Basu, Forest Hughes, Devis Tuia, and Ramesh Raskar.
\newblock Deepglobe 2018: A challenge to parse the earth through satellite
  images.
\newblock In {\em Proceedings of the IEEE Conference on Computer Vision and
  Pattern Recognition Workshops}, pages 172--181, 2018.

\bibitem{deng2009imagenet}
Jia Deng, Wei Dong, Richard Socher, Li-Jia Li, Kai Li, and Li Fei-Fei.
\newblock Imagenet: A large-scale hierarchical image database.
\newblock In {\em 2009 IEEE conference on computer vision and pattern
  recognition}, pages 248--255. Ieee, 2009.

\bibitem{handa2016scenenet}
Ankur Handa, Viorica P{\u{a}}tr{\u{a}}ucean, Simon Stent, and Roberto Cipolla.
\newblock Scenenet: An annotated model generator for indoor scene
  understanding.
\newblock In {\em 2016 IEEE International Conference on Robotics and Automation
  (ICRA)}, pages 5737--5743. IEEE, 2016.

\bibitem{he2016deep}
Kaiming He, Xiangyu Zhang, Shaoqing Ren, and Jian Sun.
\newblock Deep residual learning for image recognition.
\newblock In {\em Proceedings of the IEEE conference on computer vision and
  pattern recognition}, pages 770--778, 2016.

\bibitem{heusel2017gans}
Martin Heusel, Hubert Ramsauer, Thomas Unterthiner, Bernhard Nessler, and Sepp
  Hochreiter.
\newblock Gans trained by a two time-scale update rule converge to a local nash
  equilibrium.
\newblock {\em Advances in neural information processing systems}, 30, 2017.

\bibitem{pytorchss}
Pavel Iakubovskii.
\newblock Segmentation models pytorch.
\newblock \url{https://github.com/qubvel/segmentation_models.pytorch}, 2020.

\bibitem{ji2018fully}
Shunping Ji, Shiqing Wei, and Meng Lu.
\newblock Fully convolutional networks for multisource building extraction from
  an open aerial and satellite imagery data set.
\newblock {\em IEEE Transactions on Geoscience and Remote Sensing},
  57(1):574--586, 2018.

\bibitem{dreamtextures}
Carson Katri.
\newblock dream-textures.
\newblock \url{https://github.com/carson-katri/dream-textures}, 2022.

\bibitem{kim2018procedural}
Joon-Seok Kim, Hamdi Kavak, and Andrew Crooks.
\newblock Procedural city generation beyond game development.
\newblock {\em SIGSPATIAL Special}, 10(2):34--41, 2018.

\bibitem{kingma2014adam}
Diederik~P Kingma and Jimmy Ba.
\newblock Adam: A method for stochastic optimization.
\newblock {\em arXiv preprint arXiv:1412.6980}, 2014.

\bibitem{kong2020synthinel}
Fanjie Kong, Bohao Huang, Kyle Bradbury, and Jordan Malof.
\newblock The synthinel-1 dataset: A collection of high resolution synthetic
  overhead imagery for building segmentation.
\newblock In {\em Proceedings of the IEEE/CVF winter conference on applications
  of computer vision}, pages 1814--1823, 2020.

\bibitem{liu2020building}
Yi Liu, Chao Pang, Zongqian Zhan, Xiaomeng Zhang, and Xue Yang.
\newblock Building change detection for remote sensing images using a dual-task
  constrained deep siamese convolutional network model.
\newblock {\em IEEE Geoscience and Remote Sensing Letters}, 18(5):811--815,
  2020.

\bibitem{loshchilov2017decoupled}
Ilya Loshchilov and Frank Hutter.
\newblock Decoupled weight decay regularization.
\newblock {\em arXiv preprint arXiv:1711.05101}, 2017.

\bibitem{maggiori2017can}
Emmanuel Maggiori, Yuliya Tarabalka, Guillaume Charpiat, and Pierre Alliez.
\newblock Can semantic labeling methods generalize to any city? the inria
  aerial image labeling benchmark.
\newblock In {\em 2017 IEEE International Geoscience and Remote Sensing
  Symposium (IGARSS)}, pages 3226--3229. IEEE, 2017.

\bibitem{mayer2016large}
Nikolaus Mayer, Eddy Ilg, Philip Hausser, Philipp Fischer, Daniel Cremers,
  Alexey Dosovitskiy, and Thomas Brox.
\newblock A large dataset to train convolutional networks for disparity,
  optical flow, and scene flow estimation.
\newblock In {\em Proceedings of the IEEE conference on computer vision and
  pattern recognition}, pages 4040--4048, 2016.

\bibitem{mcinnes2018umap}
Leland McInnes, John Healy, and James Melville.
\newblock Umap: Uniform manifold approximation and projection for dimension
  reduction.
\newblock {\em arXiv preprint arXiv:1802.03426}, 2018.

\bibitem{muller2006procedural}
Pascal M{\"u}ller, Peter Wonka, Simon Haegler, Andreas Ulmer, and Luc Van~Gool.
\newblock Procedural modeling of buildings.
\newblock In {\em ACM SIGGRAPH 2006 Papers}, pages 614--623. 2006.

\bibitem{musgrave1989synthesis}
F~Kenton Musgrave, Craig~E Kolb, and Robert~S Mace.
\newblock The synthesis and rendering of eroded fractal terrains.
\newblock {\em ACM Siggraph Computer Graphics}, 23(3):41--50, 1989.

\bibitem{openai2023gpt4}
OpenAI.
\newblock Gpt-4 technical report, 2023.

\bibitem{pang2023detecting}
Chao Pang, Jiang Wu, Jian Ding, Can Song, and Gui-Song Xia.
\newblock Detecting building changes with off-nadir aerial images.
\newblock {\em Science China Information Sciences}, 66(4):1--15, 2023.

\bibitem{reyes2022syntcities}
Mario~Fuentes Reyes, Pablo d'Angelo, and Friedrich Fraundorfer.
\newblock Syntcities: A large synthetic remote sensing dataset for disparity
  estimation.
\newblock {\em IEEE Journal of Selected Topics in Applied Earth Observations
  and Remote Sensing}, 15:10087--10098, 2022.

\bibitem{richter2016playing}
Stephan~R Richter, Vibhav Vineet, Stefan Roth, and Vladlen Koltun.
\newblock Playing for data: Ground truth from computer games.
\newblock In {\em Computer Vision--ECCV 2016: 14th European Conference,
  Amsterdam, The Netherlands, October 11-14, 2016, Proceedings, Part II 14},
  pages 102--118. Springer, 2016.

\bibitem{robbins1951stochastic}
Herbert Robbins and Sutton Monro.
\newblock A stochastic approximation method.
\newblock {\em The annals of mathematical statistics}, pages 400--407, 1951.

\bibitem{rombach2022high}
Robin Rombach, Andreas Blattmann, Dominik Lorenz, Patrick Esser, and Bj{\"o}rn
  Ommer.
\newblock High-resolution image synthesis with latent diffusion models.
\newblock In {\em Proceedings of the IEEE/CVF conference on computer vision and
  pattern recognition}, pages 10684--10695, 2022.

\bibitem{ronneberger2015u}
Olaf Ronneberger, Philipp Fischer, and Thomas Brox.
\newblock U-net: Convolutional networks for biomedical image segmentation.
\newblock In {\em Medical Image Computing and Computer-Assisted
  Intervention--MICCAI 2015: 18th International Conference, Munich, Germany,
  October 5-9, 2015, Proceedings, Part III 18}, pages 234--241. Springer, 2015.

\bibitem{ros2016synthia}
German Ros, Laura Sellart, Joanna Materzynska, David Vazquez, and Antonio~M
  Lopez.
\newblock The synthia dataset: A large collection of synthetic images for
  semantic segmentation of urban scenes.
\newblock In {\em Proceedings of the IEEE conference on computer vision and
  pattern recognition}, pages 3234--3243, 2016.

\bibitem{schuhmann2022laion}
Christoph Schuhmann, Romain Beaumont, Richard Vencu, Cade Gordon, Ross
  Wightman, Mehdi Cherti, Theo Coombes, Aarush Katta, Clayton Mullis, Mitchell
  Wortsman, et~al.
\newblock Laion-5b: An open large-scale dataset for training next generation
  image-text models.
\newblock {\em arXiv preprint arXiv:2210.08402}, 2022.

\bibitem{shen2018wasserstein}
Jian Shen, Yanru Qu, Weinan Zhang, and Yong Yu.
\newblock Wasserstein distance guided representation learning for domain
  adaptation.
\newblock In {\em Proceedings of the AAAI Conference on Artificial
  Intelligence}, volume~32, 2018.

\bibitem{shermeyer2021rareplanes}
Jacob Shermeyer, Thomas Hossler, Adam Van~Etten, Daniel Hogan, Ryan Lewis, and
  Daeil Kim.
\newblock Rareplanes: Synthetic data takes flight.
\newblock In {\em Proceedings of the IEEE/CVF Winter Conference on Applications
  of Computer Vision}, pages 207--217, 2021.

\bibitem{IF}
StabilityAI.
\newblock deep-floyd.
\newblock \url{https://github.com/deep-floyd}, 2023.

\bibitem{stable2}
StabilityAI.
\newblock Stable diffusion version 2.
\newblock \url{https://github.com/Stability-AI/stablediffusion}, 2023.

\bibitem{szegedy2016rethinking}
Christian Szegedy, Vincent Vanhoucke, Sergey Ioffe, Jon Shlens, and Zbigniew
  Wojna.
\newblock Rethinking the inception architecture for computer vision.
\newblock In {\em Proceedings of the IEEE conference on computer vision and
  pattern recognition}, pages 2818--2826, 2016.

\bibitem{tan2019efficientnet}
Mingxing Tan and Quoc Le.
\newblock Efficientnet: Rethinking model scaling for convolutional neural
  networks.
\newblock In {\em International conference on machine learning}, pages
  6105--6114. PMLR, 2019.

\bibitem{vaswani2017attention}
Ashish Vaswani, Noam Shazeer, Niki Parmar, Jakob Uszkoreit, Llion Jones,
  Aidan~N Gomez, Lukasz Kaiser, and Illia Polosukhin.
\newblock Attention is all you need.
\newblock {\em Advances in neural information processing systems}, 30, 2017.

\bibitem{lovedacode}
Junjue Wang.
\newblock Loveda repository.
\newblock \url{https://github.com/Junjue-Wang/LoveDA}, 2021.

\bibitem{wang2021loveda}
Junjue Wang, Zhuo Zheng, Ailong Ma, Xiaoyan Lu, and Yanfei Zhong.
\newblock Loveda: A remote sensing land-cover dataset for domain adaptive
  semantic segmentation.
\newblock {\em arXiv preprint arXiv:2110.08733}, 2021.

\bibitem{wrenninge2018synscapes}
Magnus Wrenninge and Jonas Unger.
\newblock Synscapes: A photorealistic synthetic dataset for street scene
  parsing.
\newblock {\em arXiv preprint arXiv:1810.08705}, 2018.

\bibitem{xia2023openearthmap}
Junshi Xia, Naoto Yokoya, Bruno Adriano, and Clifford Broni-Bediako.
\newblock Openearthmap: A benchmark dataset for global high-resolution land
  cover mapping.
\newblock In {\em Proceedings of the IEEE/CVF Winter Conference on Applications
  of Computer Vision}, pages 6254--6264, 2023.

\bibitem{xie2021segformer}
Enze Xie, Wenhai Wang, Zhiding Yu, Anima Anandkumar, Jose~M Alvarez, and Ping
  Luo.
\newblock Segformer: Simple and efficient design for semantic segmentation with
  transformers.
\newblock {\em Advances in Neural Information Processing Systems},
  34:12077--12090, 2021.

\bibitem{xu2022simpl}
Yang Xu, Bohao Huang, Xiong Luo, Kyle Bradbury, and Jordan~M Malof.
\newblock Simpl: Generating synthetic overhead imagery to address custom
  zero-shot and few-shot detection problems.
\newblock {\em IEEE Journal of Selected Topics in Applied Earth Observations
  and Remote Sensing}, 15:4386--4396, 2022.

\bibitem{yang2020semantic}
Kunping Yang, Gui-Song Xia, Zicheng Liu, Bo Du, Wen Yang, Marcello Pelillo, and
  Liangpei Zhang.
\newblock Semantic change detection with asymmetric siamese networks.
\newblock {\em arXiv preprint arXiv:2010.05687}, 2020.

\bibitem{zou2020game}
Zhengxia Zou, Tianyang Shi, Wenyuan Li, Zhou Zhang, and Zhenwei Shi.
\newblock Do game data generalize well for remote sensing image segmentation?
\newblock {\em Remote Sensing}, 12(2):275, 2020.

\end{thebibliography}
